\documentclass{article} % For LaTeX2e

\usepackage{amsmath,amsfonts,bm}

\def\eqref#1{equation~\ref{#1}}
\def\1{\bm{1}}

\def\va{{\bm{a}}}
\def\vb{{\bm{b}}}

\def\mA{{\bm{A}}}

\def\mE{{\bm{E}}}

\def\mH{{\bm{H}}}

\def\mP{{\bm{P}}}

\def\mW{{\bm{W}}}
\def\mX{{\bm{X}}}

\DeclareMathAlphabet{\mathsfit}{\encodingdefault}{\sfdefault}{m}{sl}
\SetMathAlphabet{\mathsfit}{bold}{\encodingdefault}{\sfdefault}{bx}{n}

\def\gG{{\mathcal{G}}}

\def\emA{{A}}

\def\emE{{E}}

\newcommand{\R}{\mathbb{R}}

\usepackage{hyperref}
\usepackage{url}
\usepackage{graphicx}
\graphicspath{{figures/}}

\title{CoulGAT: An Experiment on Interpretability of Graph Attention Networks}

\author{Burc Gokden \\
Fromthesky Research Labs LLC\\
Oregon, USA \\
\texttt{burc@fromtheskyresearchlabs.com} \\
}

\begin{document}

\maketitle

\begin{abstract}

We present an attention mechanism inspired from definition of screened Coulomb potential. This attention mechanism was used to interpret the Graph Attention (GAT) model layers and training dataset by using a flexible and scalable framework (CoulGAT) developed for this purpose. Using CoulGAT, a forest of plain and resnet models were trained and characterized  using this attention mechanism against CHAMPS dataset. The learnable variables of the attention mechanism are used to extract node--node and node--feature interactions to define an empirical standard model for the graph structure and hidden layer. This representation of graph and hidden layers can be used as a tool to compare different models, optimize hidden layers and extract a compact definition of graph structure of the dataset.
   
\end{abstract}

\section{Introduction}

Highly irregular data structured as graphs are often encountered in social networks, e-commerce, natural language processing, knowledge databases (e.g., citation networks), quantum chemistry and molecular biology. As the relationships between nodes in a graph get more complex and irregular in size, there is a need for machine learning methods to adapt to handle these complex relationships in a more efficient way, as conventional definitions of convolution and recurrent networks are not applicable in a straight manner as they were on grid-based images or simple sequences.

Early neural networks reported to tackle graph data used recurrent network approach on a graph node's neighbors. These recurrent networks exchange information to update nodes'  states in a neighborhood until states reach equilibrium \cite{Sperduti97, Scarselli09, Micheli04} when a convergence condition is satisfied. This approach was extended to use Gated Recurrent Units allowing unrolling in time for a fixed number of steps and use of backpropagation in \cite{li2016gated}. In \cite{Dai18}, a recurrent model scalable to large graphs was explored by employing a stochastic learning algorithm using embedding representations to achieve steady state.

Convolutional networks were also successfully applied to learn from graph data. These networks learn by taking advantage of local translational invariance of data \cite{Bruna14} as compared to recurrent networks which learn causal relationships among nodes. In \cite{Bruna14}, it was shown that convolution operator for graph data can be defined by using spectral properties through graph Laplacian or by exploiting the locality spatially in a graph's neighborhood. Spectral models are domain dependent and linear transformations to derive the graph Laplacian are costly. Complexity of spectral models was reduced by using localized filters and pooling in \cite{henaff15}. The spectral model was improved in efficiency by using recursive Chebyshev polynomials as localized filters in \cite{Defferrard16}. This was followed by a simpler and scalable model using first order approximations of spectral convolutions in a 1-node neighborhood \cite{Kipf16}. Spatial models, on the other hand, learn by performing convolutions on their local neighborhood and do not rely on a well defined convolution operator derived from a graph Laplacian. An early implementation of efficient spatial graph convolution was presented in \cite{Micheli09}. In \cite{Atwood16}, a diffusion-convolution operation was defined to pass information among nodes in a probabilistic manner. Message Passing Neural Network defined a general framework for spatial convolution models in the context of quantum chemistry \cite{Gilmer17}. GraphSAGE \cite{Hamilton17} used an aggregator to generate node embeddings in an inductive manner for large graphs. Graph Attention Networks \cite{velickovic18} showed that attention is a computationally efficient mechanism that can also improve the capacity and interpretability of the model on both transductive and inductive learning tasks.

Remarkable developments in graph neural networks outlined above made these models an efficient alternative to simulations in quantum chemistry using Density Functional Theory (DFT) \cite{Gilmer17, Ryczko19, Mills17}. DFT is very successful in predicting chemical properties of materials and its results are highly interpretable due to the fact that it solves Kohn-Sham equations using energy functionals for the system. However, the computational complexity scales as $O(N^3)$ making it challenging to apply larger systems and choice of a good exchange-correlation functional can be difficult. Graph neural networks can reduce the complexity and skip calculating Kohn-Sham equations completely. Graph Attention Networks developed in \cite{velickovic18} offer an opportunity to maintain interpretability due to its flexibility in choosing an appropriate attention mechanism.

In this study, we present a scalable graph attention model framework to characterize a screened exponential Coulomb potential based attention mechanism. The framework defines building blocks for vanilla plain and resnet graph convolutional layers. Our main contribution is an attention model whose learned parameters can be used to interpret the coupling strength and interaction range for any two nodes present in a graph structure. Motivation for this interpretation was due to appearance of Coulomb potential in various contexts from describing fundamental forces of nature as in Yukawa Potential \cite{Yukawa35} to electron-electron and electron-nuclei interactions in Kohn-Sham equations \cite{Kohn65}. We use plain and resnet variations of this model on the CHAMPS dataset for predicting scalar coupling constants \cite{kaggleCHAMPS}. We show that our framework can scale up to extremely deep number of graph layers. Scalability and simplicity of models presented here are comparable to depth of Convolutional Neural Networks using skip connections for image classification \cite{He15, He16}.

In the following section, we begin first by going over attention network structure implemented in the model. We define how attention is handled within a single layer and how propagation to next layer is managed. In section \ref{Dataset}, we go over data preparation. In section \ref{Experimental Setup}, we use the basic building blocks for this network to define graph models with shallow depth and variations. We finish this section by describing deeper graph models developed using residual connections. Evaluation of these models are presented in section \ref{Results}.

\section{Model Architecture}

Our model is a graph attention network that takes as input a weighted graph $\gG=(\mX, \mA)$ where $\mX$ is $F \times N$ feature matrix for $F$ features and $N$ nodes and $\mA$ is a non-negative weighted $N\times N$ adjacency matrix. The general formalism closely follows graph attention network described in [\cite{velickovic18}. Main differences are in the way attention is obtained and adjacency matrix plays a more crucial role than being only a masking mechanism.

\subsection{Attentional Hidden Layer}

The attentional hidden layer takes as input $\mH^l \in \R^{ F \times N }$ where $l$ is layer number. First a linear transformation is applied onto input to this layer using a weight matrix $\mW^{l+1} \in \R^{F' \times F}$ and a bias term $\vb$.

\begin{equation}
\mH_{w}^{l+1}=\mW^{l+1}\mH^l+\vb       \label{eq1}
\end{equation}

This weight matrix operates on all nodes in the input to the layer. The other input for the graph, the weighted adjacency matrix $\mA$ is used to calculate attention in all layers. The weighting in $\mA$ is determined by inverse squared distance between each node $i$ and $j$. The entries for self edge and no edge are also defined by very high and very low values respectively. For this study, following formulation for $\mA$ was defined:

\begin{equation}
\emA_{ij}=
\begin{cases}
1 \times 10^3, & \text{if $i=j$}\\
1\times 10^{-5}, & \text{if edge(i,j) doesn't exist} \\
[distance(i,j)]^{-2}, & \text{if $i \ne j$}
\end{cases} \label{eq2}
\end{equation}

The adjacency matrix defines the inverse distance dependence of the Coulomb attention model and is integral to the definition of attention mechanism. Instead of relying only on attention to define the weights between neighboring nodes, we provide a priori edge weights for all possible edge connections to the attention mechanism.

All layers see the same adjacency matrix by definition for each single input graph. We make it unique by applying a pointwise exponential operation on $\mA$ with a power matrix $\mP$ which is set as learnable for each layer:

\begin{equation} \label{eq3}
{A_p}_{ij}=\dfrac {exp(\mA^{\odot\mP})_{ij}} {\displaystyle \sum_{k=1}^{N} exp(\mA^{\odot\mP})_{ik}}
\end{equation}

This modified adjacency matrix defines the exponential inverse distance part of the Coulomb potential and it is applied on $\mH^{l+1}_{w}$ as:

\begin{equation}
\mH^{l+1}_{wd}=\mH^{l+1}_{w}\mA^{T}_{p} \label{eq4}
\end{equation}

$\mH^{l+1}_{wd}$ is the intermediate hidden layer matrix weighted by modified adjacency matrix. This term reflects the effect of each node in the input on every other node as a sum weighted by the adjacency matrix elements. The matrix multiplication in eq. \ref{eq4} replaces the concatenation operation typically used for defining the attention coefficients.

The screening contribution to the Coulomb potential is achieved by using a learnable attention matrix $\va \in \R^{N \times F'} $ for finding the attention coefficients. The attention matrix elements define the screening on  the potential. Similar to inverse distance contribution, it also considers weighted contributions from all nodes:

\begin{equation}
\emE_{ij}=\frac{exp(LeakyReLU(\va^{l+1}\mH^{l+1}_{wd})_{ij})} { \displaystyle \sum_{k=1}^{N} exp(LeakyReLU(\va^{l+1}\mH^{l+1}_{wd})_{ik})} \label{eq5}
\end{equation}

Finally, next hidden layer is formed by multiplying attention coefficients with the input and applying the activation  layer.

\begin{equation}
\mH_{single}^{l+1}=\sigma(\mH^{l+1}_{w}\mE^T) \label{eq6}
\end{equation}

The stability and capacity of each layer is improved by utilizing $K$ heads for each node, therefore above operation was performed separately $K$ times with $K$ weight and attention factors and the $K$ output matrices are concatenated to form a $\R^{KF' \times N}$ matrix as output to next layer:

\begin{equation}
\mH^{l+1}=\overset {K} {\underset {k=1} {\Big\Vert}}  \mH^{l+1,k}_{single} \label{eq7}
\end{equation}

After output of last attention layer is flattened to a 1-D vector, this is fed into a 3-layer fully connected layer consisting of two $K \times F$ dense layers and final layer is equal to the number of classify labels.

\subsection{Interpretability of graph attention function}

Attention between nodes is modeled after screened Coulomb potential which is a potential with an exponential damping term. The damping term defines the range of the interaction between two particles. It is of the general form:

\begin{equation}
C\frac{e^{-Mr}}{r} \label{eq8}
\end{equation}

Where $C$ is the magnitude scaling constant and $M$ scales the range of interaction. For $M=0$, it reduces to the form of electromagnetic Coulomb potential. In our model, after expanding the inner term in eq. \ref{eq5}, this potential is approximated as:

\begin{equation}
\va^{l+1}[(\mW^{l+1}\mH^l+\vb)\mA_p^T] \label{eq9}
\end{equation}

The attention matrix $\va$ and adjacency matrix $\mA_p$ approximate the screening coefficients and inverse distance effect of the Coulomb potential. Different from eq. \ref{eq8}, the exponential is defined on the inverse distance in the proposed attention function instead of $e$. The attention mechanism takes into account of linear combination of all neighboring nodes weighted by using a fully learnable attention matrix $\va$ and a priori provided weighted adjacency matrix $\mA_p$. The adjacency matrix was defined to include self interactions (and non-existing interactions) such that exchange-correlation contribution of the DFT calculations can be part of the learning process.

This attention model learns the range, magnitude and distance power of interaction forces between each node in a graph. Since we did not put any limitations on the attention mechanism such as immutable constants and specific charge carriers, this model, in principle, may learn any short and long range interactions between two nodes regardless of the nature of actual (or artificially defined) forces between them. This makes the model very flexible and applicable to wide range of data that can be represented as graph, of which a weighted distance between (many) nodes can be defined. Consequently, interaction between graph nodes can be classified based on their coupling strength and interaction range by reviewing the attention coefficient values.

\section{Dataset} \label{Dataset}
We use a quantum chemistry dataset from CHAMPS institute made available in \cite{kaggleCHAMPS}. This dataset lists scalar coupling constants between pairs of atoms in a molecule based on their interaction type. Scalar coupling constants are magnetic interactions between atoms that can be used in NMR to understand a molecule's structure. The dataset also provides the $(x, y, z)$ coordinates for each atom in a molecule for train and test data.

In this dataset each scalar coupling value is provided as a single entry. The data needs preprocessing to define graphs for each molecule within the dataset. Each molecule is a graph that can be a feature input $\mX$ to the graph attention model accompanied by their weighted adjacency matrix $\mA$. In the dataset, the atom\_index\_0 and scalar coupling type are combined to form a 2--tuple. These 2--tuples are then converted to a one-hot vector for each entry. This creates a sparse vector for each node. This representation is not best for accuracy and an embedding could improve model performance. We use this sparse representation for its convenience in demonstrating the interpretability of attention mechanism. The molecules which have same atom\_index\_1 within same molecule are summed row-wise to accumulate feature vectors for each atom in a molecule. The feature number $F$ after one-hot vector expansion was 211 and maximum number of atoms a molecule can have was found to be 29 in this dataset. For each molecule, data is processed to extract a sparse feature matrix $\mX$ with size $F \times N$, a flattened scalar coupling matrix as y labels of size $FN$. The adjacency matrix $\mA$ is also processed to find the distance between each atom in a molecule as defined in eq. \ref{eq2}. Although $\mX$ is sparse, $\mA$ is not and its values are non-negative real. The preprocessing was done for both train and test data available in \cite{kaggleCHAMPS}. The models were trained on the train data after splitting train/validation/test sets in  70/15/15 split ratio for 85003 molecules processed.

\section{Experimental Setup} \label{Experimental Setup}
By using the proposed coulomb attention model, deep learning models with varying architectures, filter sizes and hyperparameter settings are built and characterized. We first discuss the shallower plain models which only use attention layers and fully connected layers for prediction with large weight matrices. These models establish a stable baseline for deeper models. We then show how these shallow model components can be used to scale to much deeper models with smaller weight matrices for each layer and by using residual blocks. All models were built using CoulGAT framework\footnote{Implementation of CoulGAT can be found at https://github.com/burcgokden/CoulGAT-Graph-Attention-Interpretability.} that was implemented with Tensorflow \cite{tfwp2015}. 

In the next section, we present plain graph attention models and resnet graph attention models with identity connections.

\subsection{Plain Models}

\begin{figure}[h!]
\centering
\begin{minipage}[b]{0.4\textwidth}
\centering
 \includegraphics[width=\textwidth]{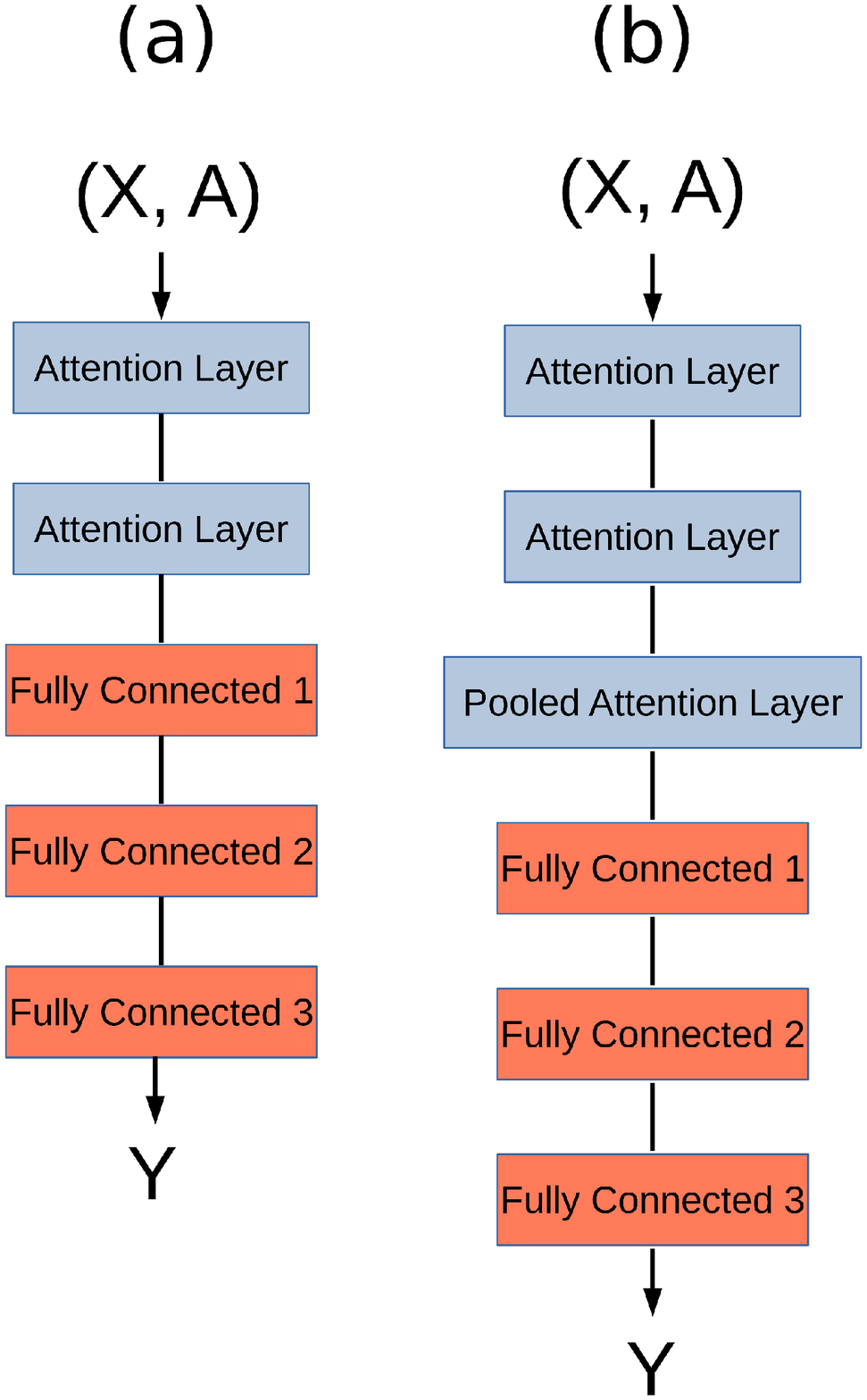}
\end{minipage}
\hfill
\begin{minipage}[b]{0.8\textwidth}
\centering
 \includegraphics[width=\textwidth]{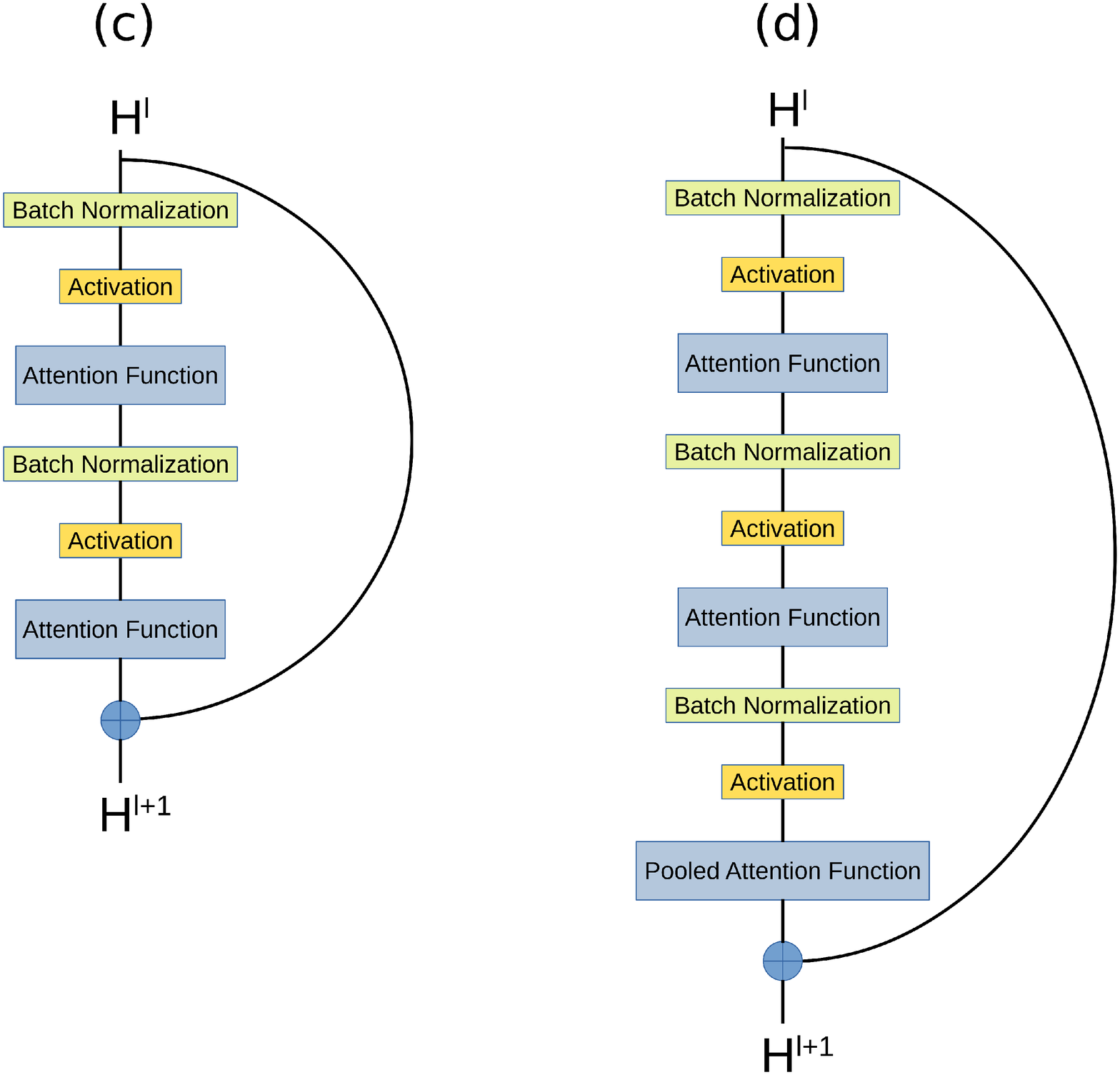}
\end{minipage}
\caption{Basic building blocks of graph attention layers described in this study. (a) 2--layer attention plaing attention layer (b) 3--layer plain attention layer with pooled attention as last layer. (c) Resnet block for 2--unit attention layer with preactivation. (d) Resnet block for 3--unit attention layer with preactivation and pooling in last layer.}
\label{fig1}
\end{figure}

The structure of a plain model is shown in Fig. \ref{fig1}(a,b). The model accepts a batch of input graphs $\gG=(\mX, \mA)$ for each epoch. Batch size was kept at 128 and train/validation set size was 59500/12750 (molecules, or graphs).  The input set $(\mX,\mA)$ first goes through at least two graph attention layers. The output of graph attention layers was flattened and fed into a 3--stage dense layer which was scaled by number of $K$ heads and input feature count $F$. The last dense layer outputs $NF$ real regression values for each $(\mX, \mA)$ predicting the scalar coupling values for each atom pair and coupling type within that model. Since all molecules do not have all $N$ atoms in their graph, the output is a sparse vector.

Model was regularized by employing drop-out \cite{Srivastava14} keep probabilities of 0.8 on input matrix $\mH_{w}$ to a hidden graph attention layer and the resulting attention coefficients in $\mE$ within each hidden layer. The dense layers were also regularized with L2, and the regularization scale was kept at 0.0001 for all training sessions. Adam optimizer was used with a learning rate of 0.001. Above hyperparameters were used when MSE was used as the loss function. All elements of the sparse output vector was used in MSE loss calculation, so this version also estimated the zeros.

For log(MAE) based loss, the hyperparameters changed as follows: dropout was set to 1.0, and L2 regularization scaled by a higher value of 0.0005. Learning rate was kept same. The SCCLMAE measure used in this study takes only log MAE of the non-zero elements in the output vector and it is more pessimistic than the LMAE defined in \cite{kaggleCHAMPS} since it averages over 211 features. The zero elements of sparse output vector was kept as free variables for this loss measure.

Unless otherwise noted ReLU was used for layer activations and LeakyReLU was used for attention function. Weight, attention and power variables were initialized using glorot initialization and biases were initialized to zeros.
 
Table \ref{table11} shows the variations of graph attention models trained. These models are trained without batch normalization or identity connections, hence called in this paper as plain. Hidden layer feature sizes were varied from $2F$ to $F$, for $F=211$. The layer depth for plain models trained had 2 or 3 graph attention layers. The head was kept at K=5 for most models and $K=2$ was also trained. High drop-out rate of 0.5 and optimization without learnable power variables was also tried.

In some models last graph attention layer employed a pooling approach where concatenation operation in eq. \ref{eq7} was replaced with sum divided by number of heads, $K$:

\begin{equation}
\mH^{l+1}=\frac{1}{K} \sum_{k=1}^{K} \mH^{l+1,k}_{single} \label{eq10}
\end{equation}

These averaging layers were also coupled with concatenating layer to form blocks repeated as defined in Table \ref{table11}.

\clearpage

\subsection{Resnet Models}

Resnet models used the same hyperparameters for batch size, optimization and regularization depending on loss measure. Batch normalization momentum was set to 0.9 and 0.99 for MSE and SCCLMAE losses, respectively. To form deeper models, plain models are modified to include batch normalization before layer activation (preactivation), and identity connections \cite{He15, He16}. Residual blocks that are comprised of graph attention layers terminated with and without a final average-pooling graph layer as in Fig \ref{fig1}(c,d).

2--unit and 3--unit resnet blocks are used to form deeper learning models as summarized in Table \ref{table21}. The feature size on hidden layers are changed to 50, 105 and 211 to reach deep learning models having 140, 30 and 12 hidden graph attention layers, respectively. For each resnet model, first layer is a single graph attention layer or a block attention layer (with averaging at the end) without identity connection originating from $\mX$. This first layer is terminated with a batch normalization+ReLU and its output serves as the first identity connection for the model. With resnet structure presented here, models that show a compromise between number of $K$ heads and layer depth are also implemented featuring a 40 hidden layer model with 10 heads for hidden feature size of 50. 

\section{Results} \label{Results}

In this section, we present results from a variety of plain and resnet models implemented and run for 100 epochs and 120 epochs for MSE, and 200 epochs for SCCLMAE loss with early stopping unless noted otherwise. Here, we evaluated these models up to the number of epochs trained to show the capacity, and extensibility of this graph model with screened coulomb potential attention. The hyperparameters we chose could be optimized for each model defined here to get a better performance. We compared all models with similar hyperparameters for this study as scanning all optimum hyperparameter values for each model is costly. We believe these results will present a useful baseline to create optimized and scalable model architectures for other datasets that can utilize the proposed approach.

\begin{table}[ht]
\caption{Plain Model Architectures}
\label{table11}
\centering
\resizebox{0.9\textwidth}{!}{

\begin{tabular}{c c c c c c }
\hline 
Model Name & \# Heads & GAT Feature & \# GAT Layers & Pool Layer & Model Type \\ 
\hline 
Plain \#1 & 5 & 422 & 2 & No & Model 1  \\ 
\hline 
Plain \#2 & 5 & 422 & 2 & Yes & Model 1  \\ 
\hline 
Plain \#3 & 5 & 422 & 3 & No & Model 1 \\ 
\hline 
Plain \#4 & 5 & 422 & 3 & Yes & Model 1  \\ 
\hline 
Plain \#5 & 5 & 211 & 3 & Yes & Model 1 \\ 
\hline 
Plain \#6 & 2 & 422 & 3 & Yes & Model 1 \\ 
\hline 
\begin{tabular}{@{}c@{}} Plain \#7 \\ (dropout 0.5) \end{tabular} & 5 & 422 & 3 & Yes & Model 1 \\ 
\hline 
\begin{tabular}{@{}c@{}} Plain \#8 \\ (no pwr) \end{tabular}  & 5 & 422 & 3 & Yes & Model 1 \\ 
\hline 
\end{tabular}}

\bigskip

\caption{Plain Model Loss Values}
\label{table12}
\resizebox{0.8\textwidth}{!}{
\begin{tabular}{c c c c c}
\hline
Model Name & \multicolumn{2}{c} {MSE}  &  \multicolumn{2}{c} {SCCLMAE}  \\
\hline
 & Min Train Loss & Min Val Loss & Min Train Loss & Min Val Loss \\
\hline
Plain \#1  & 0.235 & 0.256 & 0.638 & 0.811 \\ 
\hline 
Plain \#2 & 0.236 & 0.262 & 0.709 & 0.868  \\ 
\hline 
Plain \#3 & 0.214 & 0.247 & 0.654 & 0.873  \\ 
\hline 
Plain \#4 & 0.201 & 0.237 & 0.705 & 0.903  \\ 
\hline 
Plain \#5 & 0.222 & 0.254 & 0.773 & 0.935  \\ 
\hline 
Plain \#6 & 0.256 & 0.282 & 0.709 & 0.873  \\
\hline 
\begin{tabular}{@{}c@{}} Plain \#7 \\ (dropout 0.5) \end{tabular} & 0.251 & 0.273 & n/a & n/a \\ 
\hline 
\begin{tabular}{@{}c@{}} Plain \#8 \\ (no pwr) \end{tabular}  & 0.193 & 0.222 & n/a & n/a \\ 
\hline 
\end{tabular} }

\end{table}

\subsection{Plain Model Evaluation}

The minimum training and validation losses are summarized for each model in Table \ref{table12} for plain models. Layer parameters are shown in Table \ref{table11}.

\textbf{MSE Loss.} MSE loss tries to predict all entries in the sparse  output vector, thus being averaged over full vector length. We start with a 2 layer graph attention model in plain \#1 and replace the last attention layer in plain \#2 with pooling. The MSE loss did not show a significant change with pooling. For the 3 layer models, plain \#3 and plain \#4, we see slightly improved loss with pooling. Using plain \#3 as a base, reducing capacity in \#5 and \#6 by reducing filter size by half and head number by 3 increased the MSE loss overall. We also tried higher dropout in plain \#7 (150 epochs) and no learnable power coefficients in plain \#8. Higher dropout increased the MSE loss. The no power coefficient model has improved loss, which may indicate glorot initialization may not be the optimum choice for initialization. 

\textbf{SCCLMAE Loss.} When loss measure is changed to SCCLMAE, the dropout was disabled and L2 regularization was increased to 0.0005 for all models to reduce overfitting. Dropout was disabled to set a baseline for resnet models with SCCLMAE loss due to detrimental effects of dropout on batch normalization \cite{LiBN18}. SCCLMAE loss averages only on observed values of the dataset in the sparse output vector. The pooling layer implementations in plain \#2 and \#4 had consistently higher losses. Reducing feature size made the loss worse but reducing K by 3 units did not prove as detrimental in plain \#6. For all models, the train and validation curves showed a larger gap for this loss measure. 

The SCCLMAE and MSE loss curves for Plain \#3 and Plain \#4 are shown in Fig. \ref{fig2}. Loss curves for other models can be found in Appendix. For all models, test data split loss was in good agreement with minimum validation loss when evaluated using model obtained by early stopping.

\begin{figure}[h!]
\centering
\begin{minipage}[b]{0.6\textwidth}
\centering
 \includegraphics[width=\textwidth]{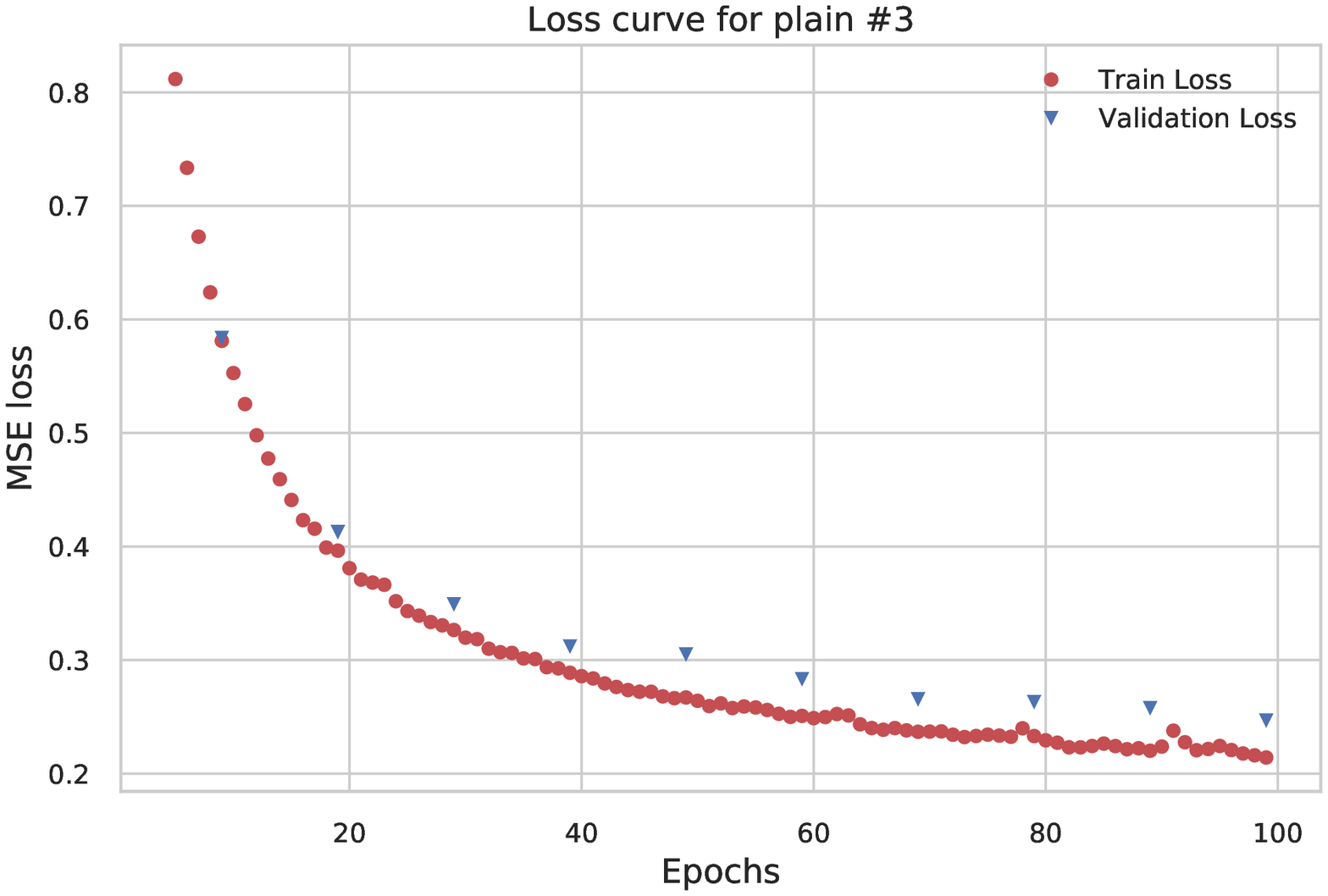}
\end{minipage}
\hfill
\begin{minipage}[b]{0.6\textwidth}
\centering
 \includegraphics[width=\textwidth]{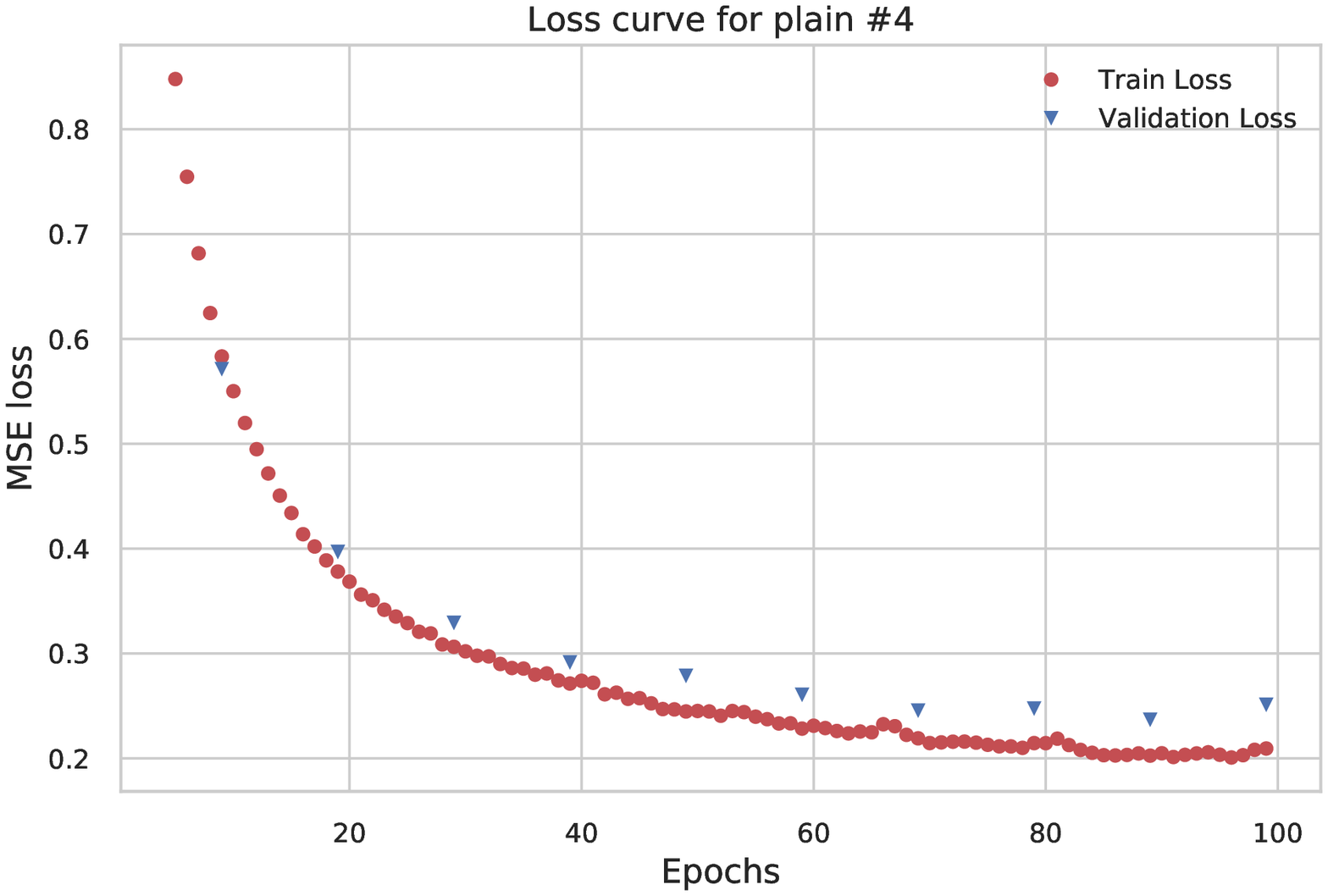}
\end{minipage}
\hfill
\begin{minipage}[b]{0.6\textwidth}
\centering
 \includegraphics[width=\textwidth]{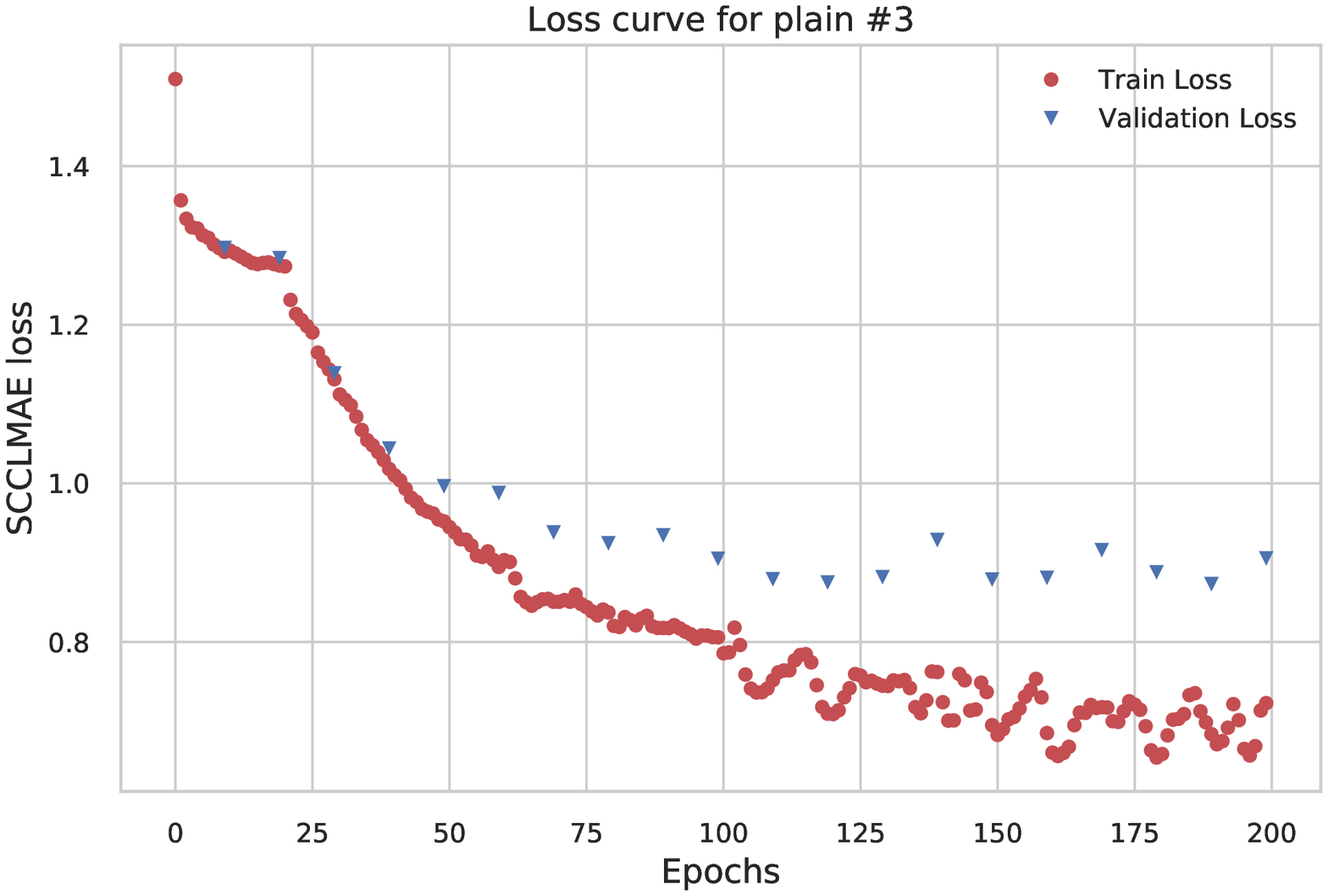}
\end{minipage}
\hfill
\begin{minipage}[b]{0.6\textwidth}
\centering
 \includegraphics[width=\textwidth]{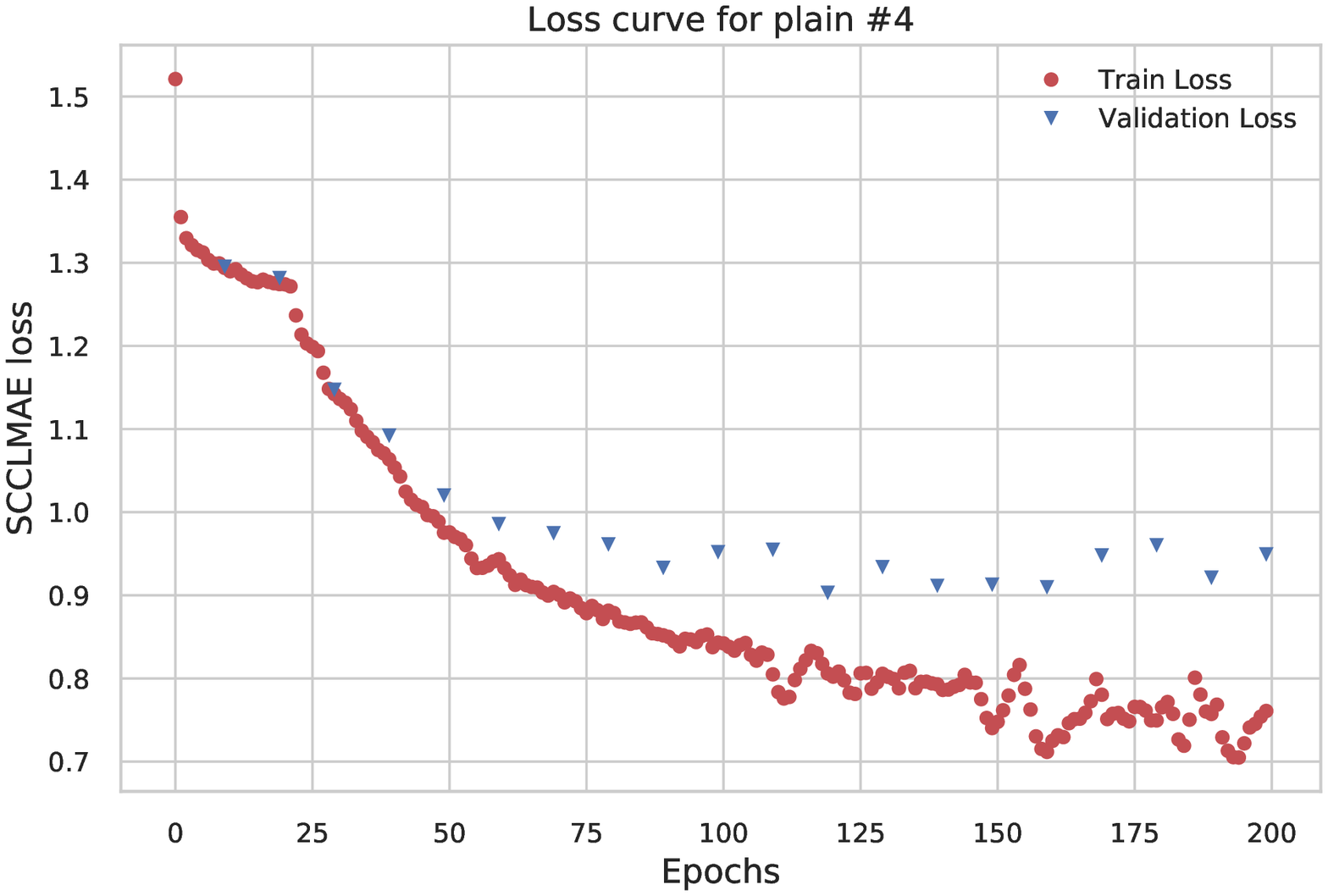}
\end{minipage}
\caption{Loss curves for plain \#3 and plain \#4 models detailed in Table \ref{table11}.}
\label{fig2}
\end{figure}
\clearpage

\subsection{Resnet Model Evaluation}
The resnet models trained for this study are summarized in Table \ref{table21} and losses are shown in Table \ref{table22}.

\textbf{MSE Loss.} The MSE train and validation loss curves converge smoothly for resnet models that ran for 120 epochs. This is attributed to the fact that loss is averaged over all elements of sparse vector, but this smoothing effect in loss definition also limits the expressive power of the model.  Pooling within a res--block or as a final layer is reflected as higher validation losses in resnet \#1, \#2 and \#4. Especially the pooling layer as a final block increase the baseline for minimum loss. For MSE loss, deeper models in resnet \#6 and \#7 does not improve loss and even degraded in resnet \#6. As a future work, the hyperparameters could be optimized to improve performance of deeper residual models.

\textbf{SCCLMAE Loss.} SCCLMAE loss curves fluctuate more, especially for the validation loss. This is attributed to averaging of SCCLMAE loss only for observed values which are few in number. There is a gap between training and validation curves which seem to level in a band around $\sim 0.8$. Pooling effects are more clear in resnet \#1, \#2 and \#4 with SCCLMAE loss. Pooling increases the minimum loss for train and validation. This is more evident in resnet \#4 which also has a narrower gap between train and validation losses. Resnet \#6 has the deepest number of layers up to 140 and scaling up head number and sacrificing depth also works well in resnet \#7. Deepest models (\#5,\#6,\#7) also show bigger gap between train and validation curves which can benefit from optimizing the hyperparameters further. 

\begin{table}[ht]
\caption{Resnet Model Architectures}
\label{table21}
\centering
\resizebox{1\textwidth}{!}{

\begin{tabular}{c c c c c c c }
\hline 
Model Name & \# Heads/Block & Feature/Block & \# Block Layers & Pool in Block & Pool at the end & Model Type\\ 
\hline 
Resnet \#1 & [5,5,5] & [211,211,211] & 5 & Yes & No & Model 4   \\ 
\hline 
Resnet \#2 & [5,5] & [211,211] & 8 & Yes & No & Model 4 \\ 
\hline 
Resnet \#3 & [5,5] & [211,211] & 6 & No & No & Model 2 \\ 
\hline 
Resnet \#4 & [5,5] & [211,211] & 6 & No & Yes & Model 2  \\ 
\hline 
Resnet \#5 & [5,5] & [105,105] & 15 & No & No & Model 2 \\ 
\hline 
Resnet \#6 & [5,5] & [50,50] & 70 & No & No & Model 2 \\ 
\hline 
Resnet \#7 & [5,5] & [50,50] & 20 & No & No & Model 2 \\ 
\hline 
\end{tabular}}

\bigskip

\caption{Resnet Model Loss Values}
\label{table22}
\resizebox{0.8\textwidth}{!}{
\begin{tabular}{c c c c c}
\hline
Model Name & \multicolumn{2}{c} {MSE}  &  \multicolumn{2}{c} {SCCLMAE}  \\
\hline
 & Min Train Loss & Min Val Loss & Min Train Loss & Min Val Loss \\
\hline
Resnet \#1  & 0.199 & 0.252 & 0.706 & 0.831 \\ 
\hline 
Resnet \#2 & 0.220 & 0.272 & 0.673 & 0.808  \\ 
\hline 
Resnet \#3 &  0.189 & 0.241 & 0.629 & 0.783  \\ 
\hline 
Resnet \#4 &  0.241 & 0.286 & 0.756 & 0.858  \\ 
\hline 
Resnet \#5 & 0.201 & 0.259 & 0.597 & 0.807  \\ 
\hline 
Resnet \#6 &  0.251 & 0.305 & 0.615 & 0.802  \\
\hline 
Resnet \#7 & 0.202 & 0.257 & 0.605 & 0.782  \\
\hline 
\end{tabular} }

\end{table}

\begin{figure}[!h]
\centering
\begin{minipage}[b]{0.6\textwidth}
 \centering
 \includegraphics[width=\textwidth]{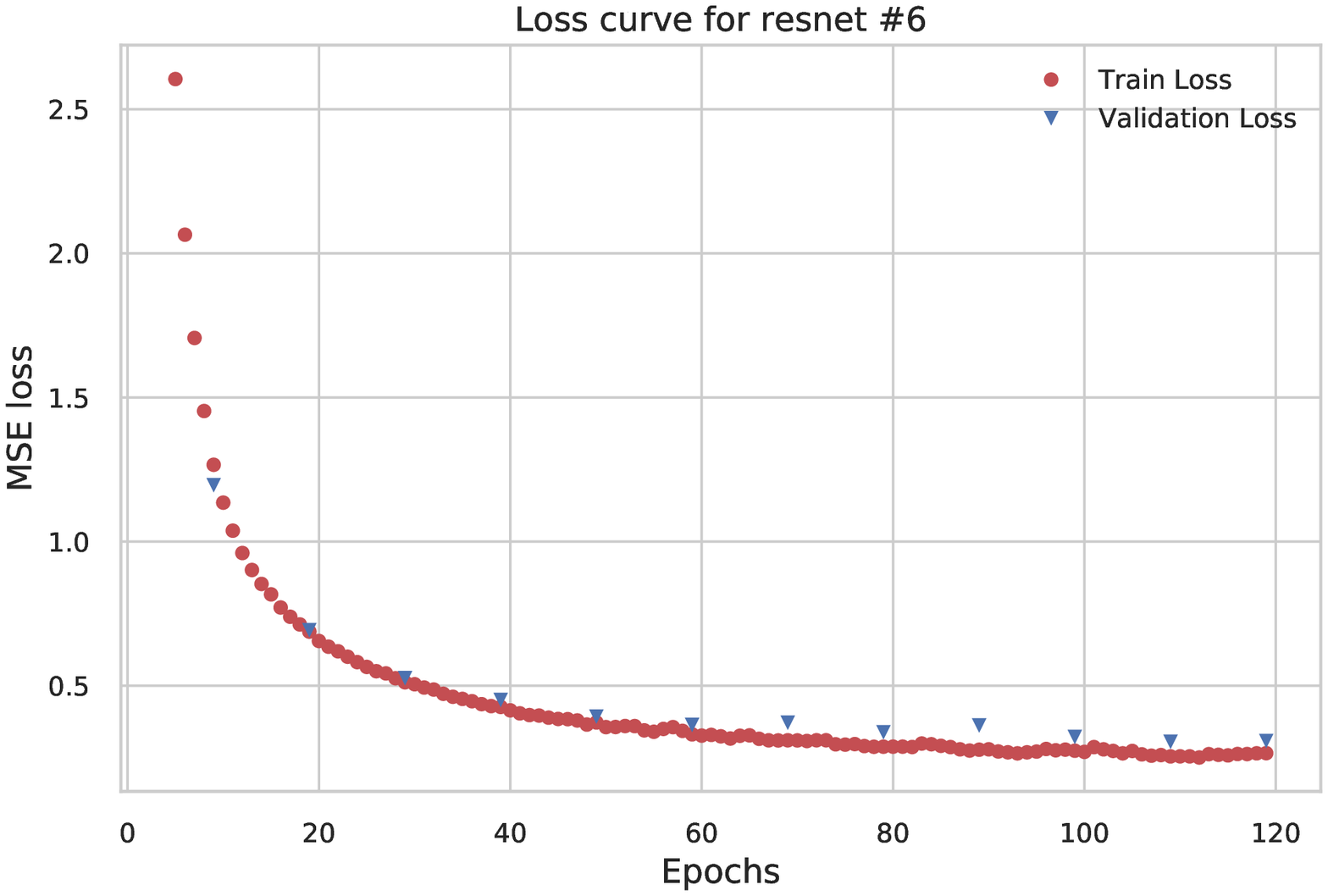}
\end{minipage}
\hfill
\begin{minipage}[b]{0.6\textwidth}
 \centering
 \includegraphics[width=\textwidth]{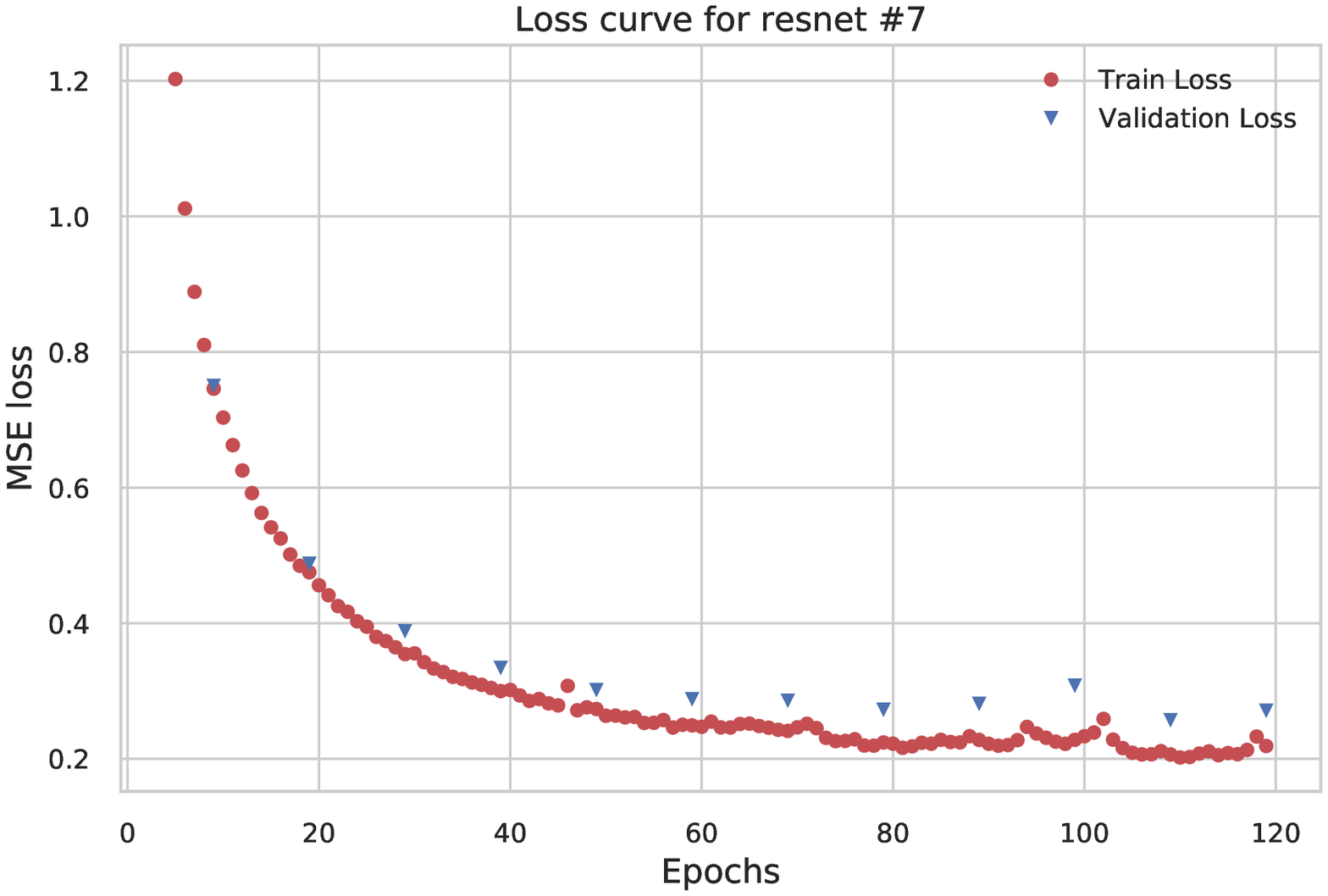}
\end{minipage}
\hfill
\begin{minipage}[b]{0.6\textwidth}
 \centering
 \includegraphics[width=\textwidth]{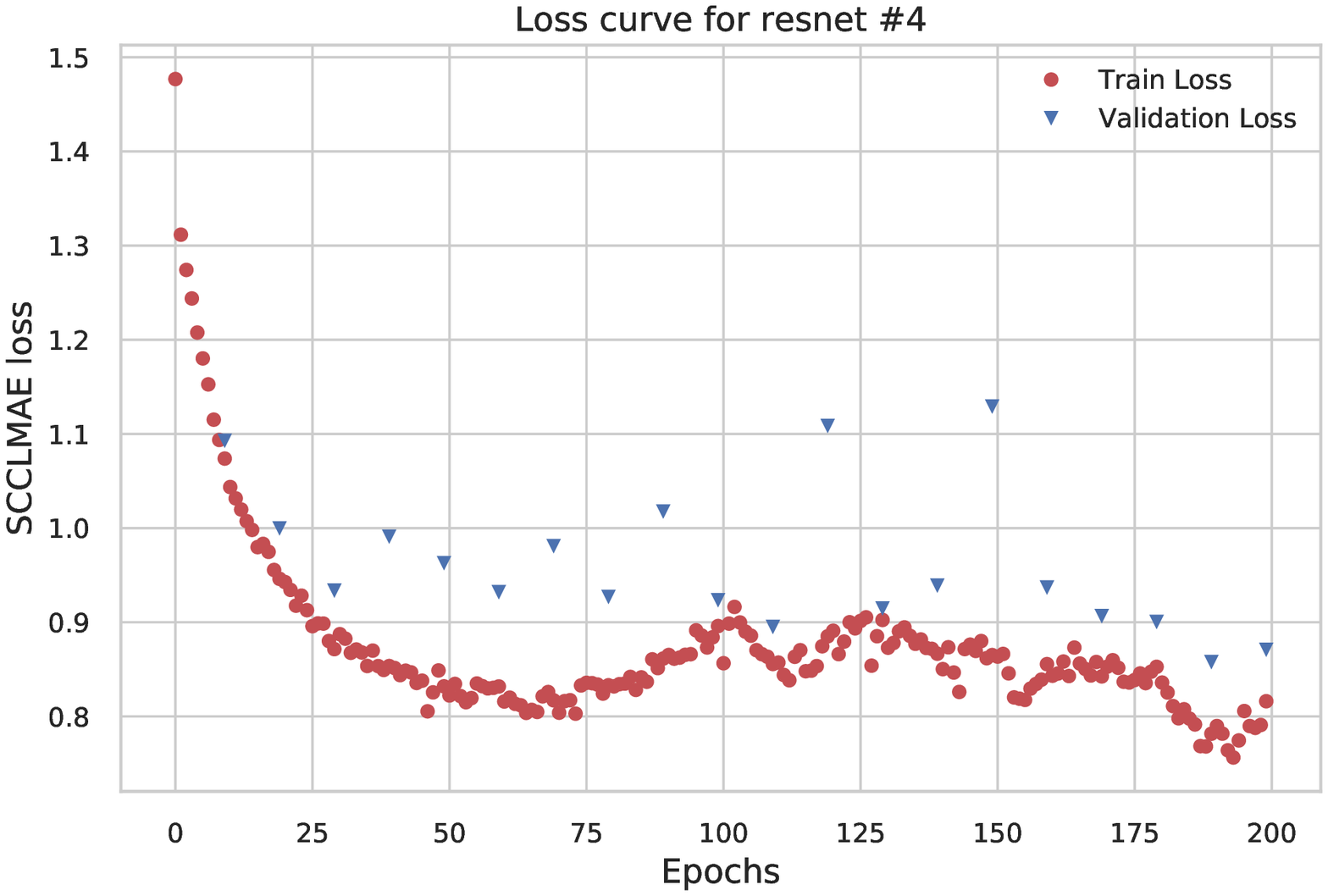}
\end{minipage}
\hfill
\begin{minipage}[b]{0.6\textwidth}
 \centering
 \includegraphics[width=\textwidth]{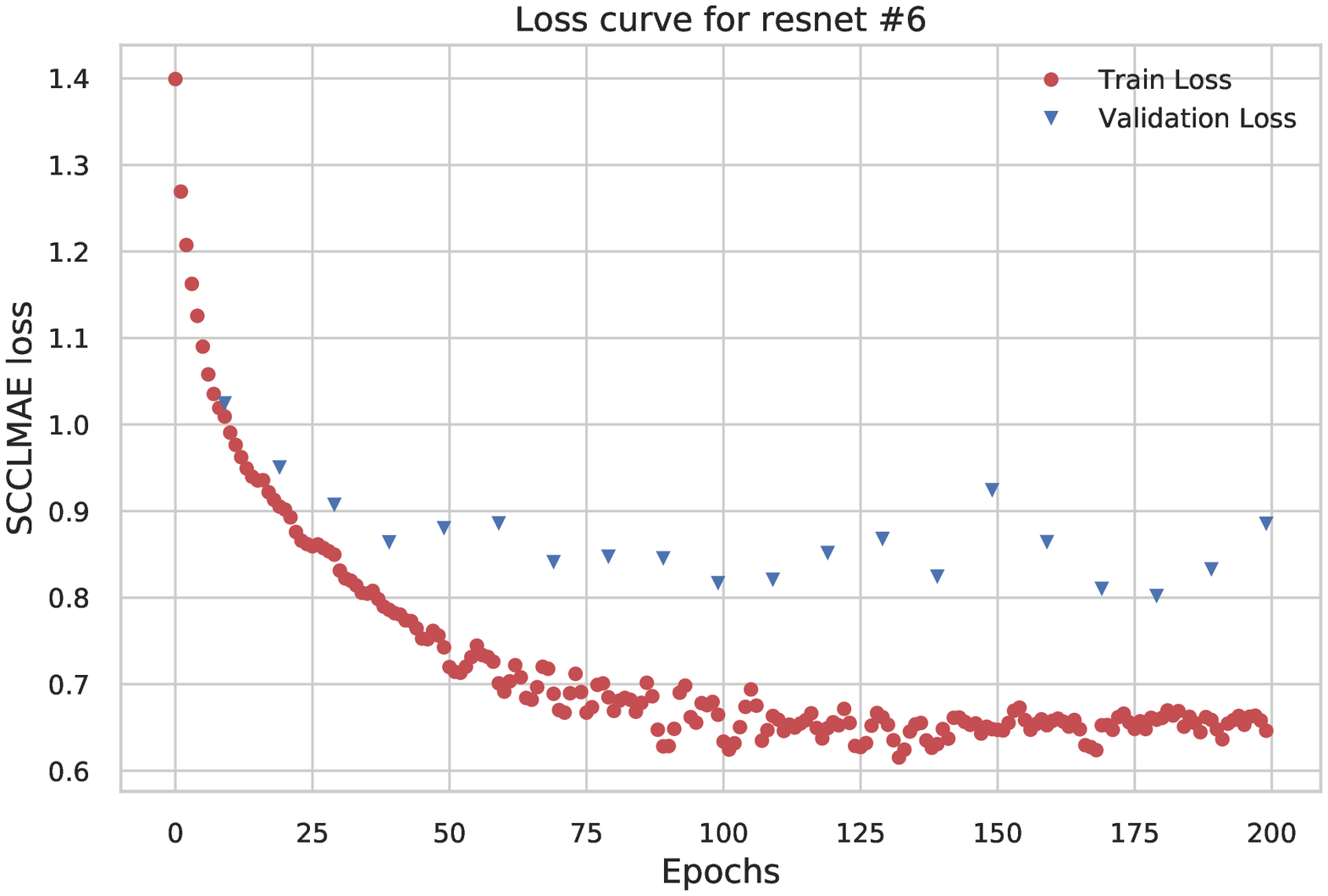}
\end{minipage}
\caption{Loss curves for resnet \#6, \#7 (MSE) and resnet \#4, \#6 (SCCLMAE) as detailed in Table \ref{table21}.}
\label{fig3}
\end{figure}

Fig. \ref{fig3} shows MSE loss curve for Resnet \#6, \#7 and SCCLMAE loss curves for Resnet \#4, \#6. Other loss curves can be found in Appendix.

\clearpage

\begin{figure}[!h]
\centering
\begin{minipage}[b]{0.6\textwidth}
 \centering
 \includegraphics[width=\textwidth]{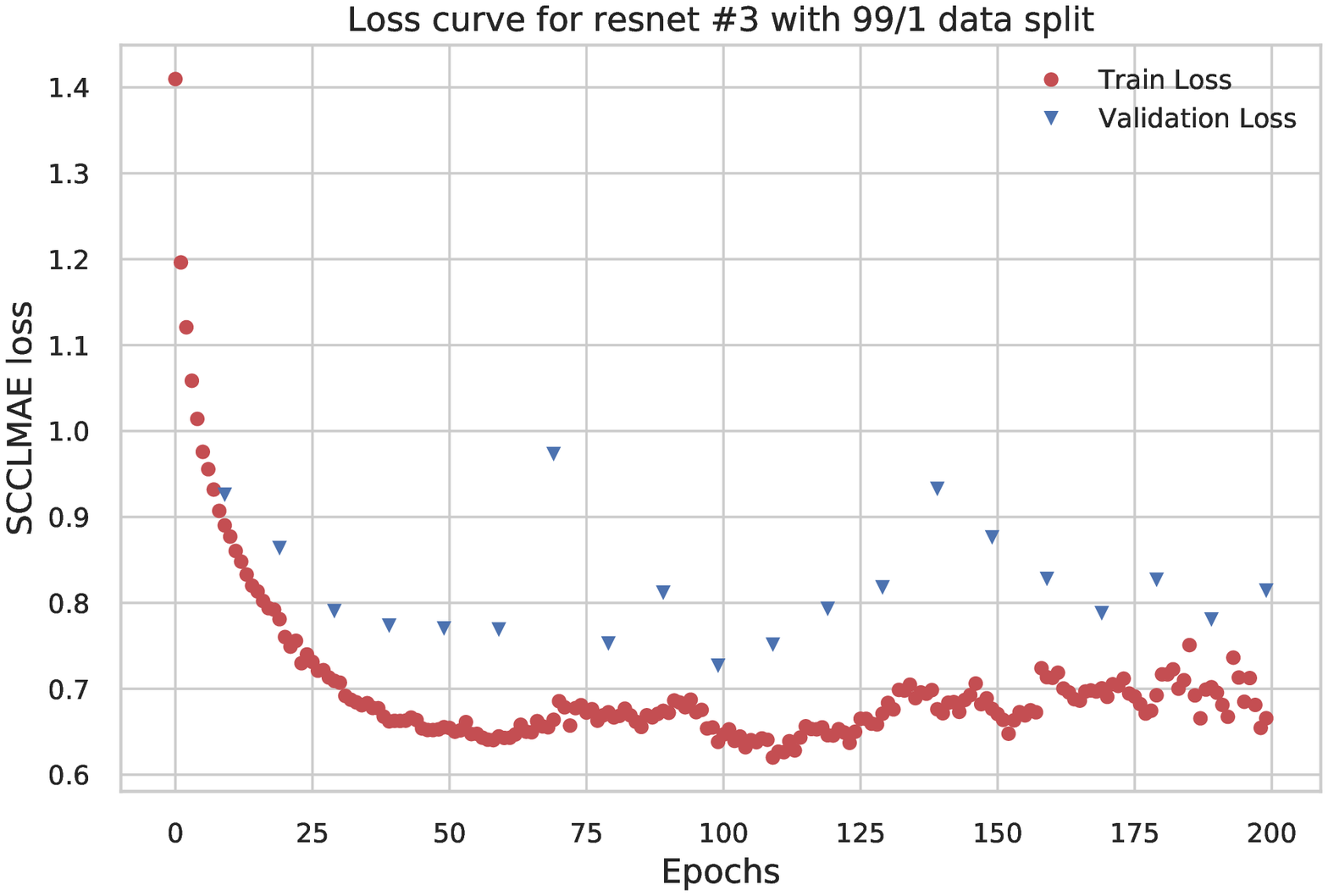}
\end{minipage}
\hfill
\begin{minipage}[b]{0.6\textwidth}
 \centering
 \includegraphics[width=\textwidth]{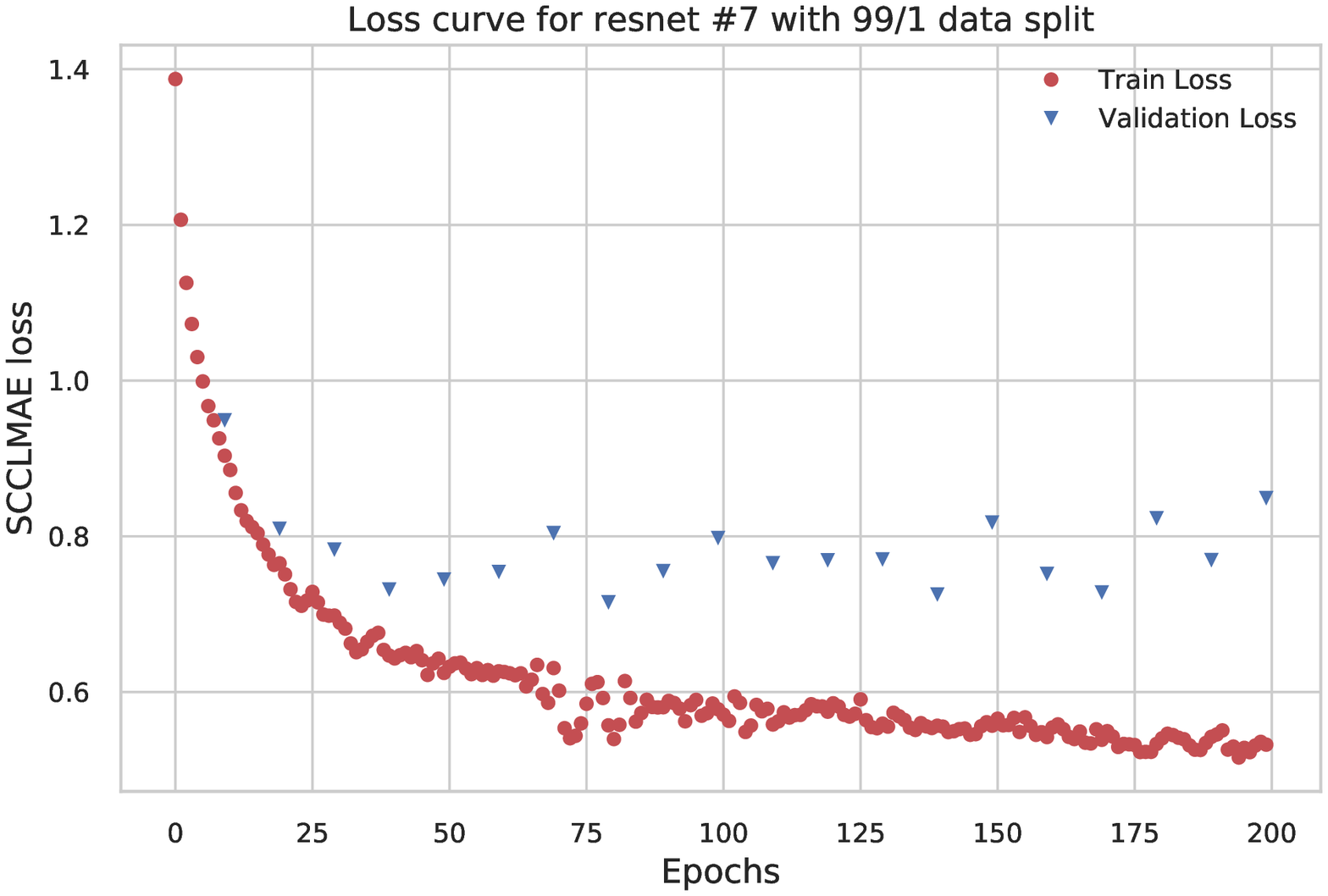}
\end{minipage}
\caption{Loss curves for Resnet \#3 (top) and \#7 (bottom) trained with 99/1 dataset split ratio.}
\label{fig6}
\end{figure}

For evaluating the Kaggle score with test data from \cite{kaggleCHAMPS}, resnet \#3 and resnet \#7 were trained for 200 epochs with 84k molecules from training set and 1k molecules were used for validation(99/1 split ratio). The SCCLMAE validation losses were 0.727 and 0.715 for resnet \#3 and Resnet \#7. The Kaggle LMAE test score of these models were 0.586(resnet \#3) and 0.596(resnet \#7). Loss curves for these models are shown in Fig. \ref{fig6}.

\begin{figure}[!h]
\centering
 \includegraphics[width=0.6\textwidth]{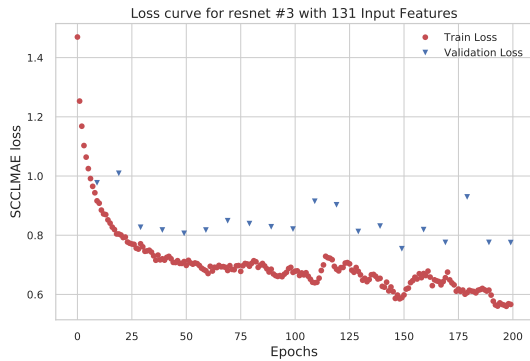}
\caption{Loss curve for Resnet \#3 trained with 131 feature dataset for 200 epochs.}
\label{fig7}
\end{figure}

We also evaluated dataset preprocessed using (atom\_index\_1, type) combination as a one-hot vector representation of input features. The resulting input feature size for this representation was 131. We trained 131--feature dataset on resnet \#3 using splitting dataset with 70/15/15 split ratio.  Loss curve is shown in Fig. \ref{fig7}. The minimum validation loss was 0.754 and minimum train loss was 0.560, outperforming resnet \#3 results in Table \ref{table22}.

\subsection{Evaluation of Learnable Attention Parameters}

The learnable attention coefficients $\va$ and $\mP$ of the attention function in each layer contains valuable information for generalizing node--to--node and node--to--feature interactions. As it was stated earlier, $\va$ defines a set of coupling strengths from atomic\_index\_1 to a tuple (atomic\_index\_0, type) and $\mP$ defines a set of interaction ranges between each atom(node) in a molecule(graph) and it is a power matrix applied to the weighted adjacency matrix consisting of inverse square of distance values ($ \sim \frac 1 {r^2}$). These two learnable parameters provide an easy way to understand how a hidden layer interprets the graph structure it was trained on. 

\begin{figure}[!h]
\centering
\begin{minipage}[b]{0.7\textwidth}
 \centering
 \includegraphics[width=\textwidth]{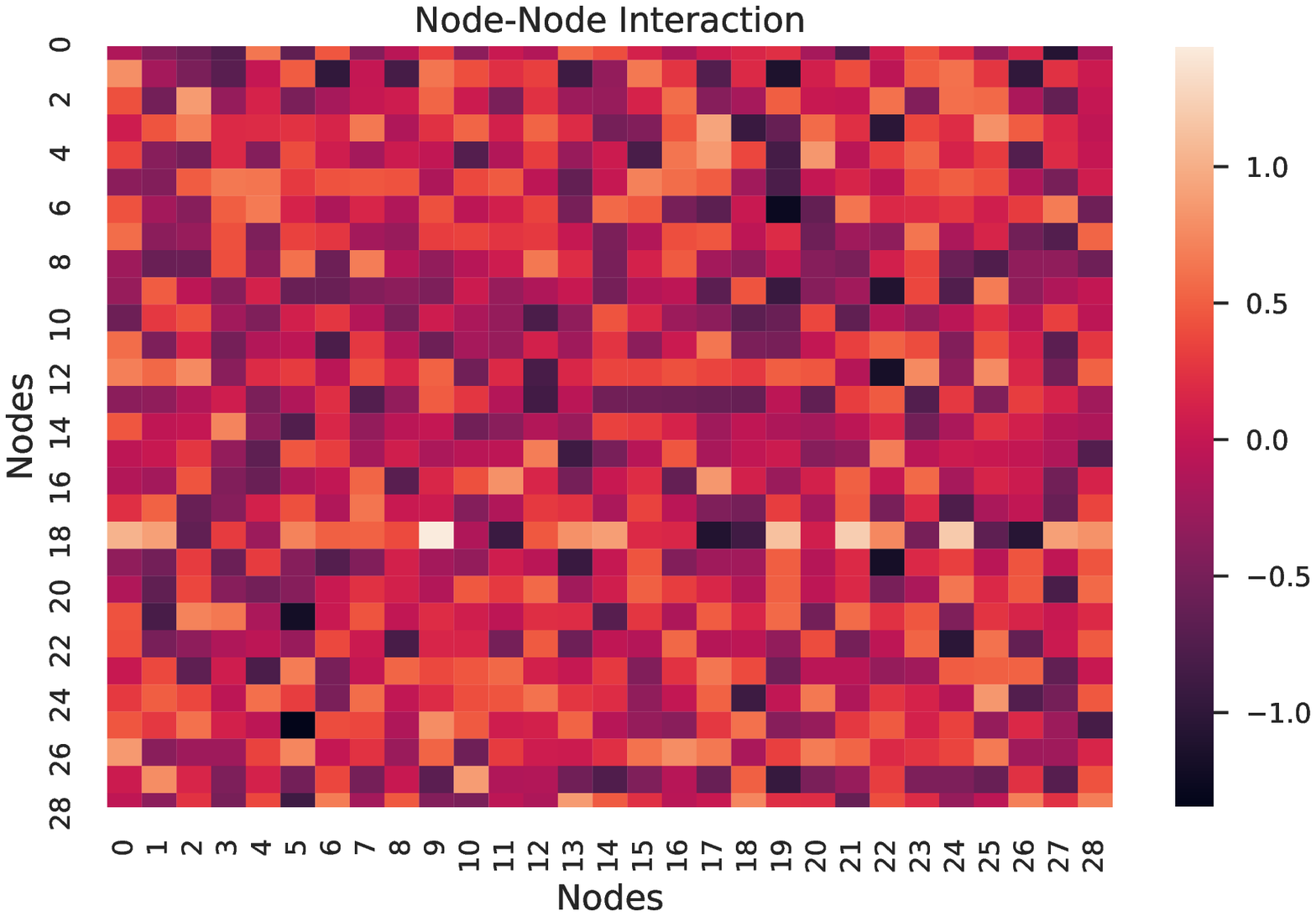}
\end{minipage}
\hfill
\begin{minipage}[b]{0.7\textwidth}
 \centering
 \includegraphics[width=\textwidth]{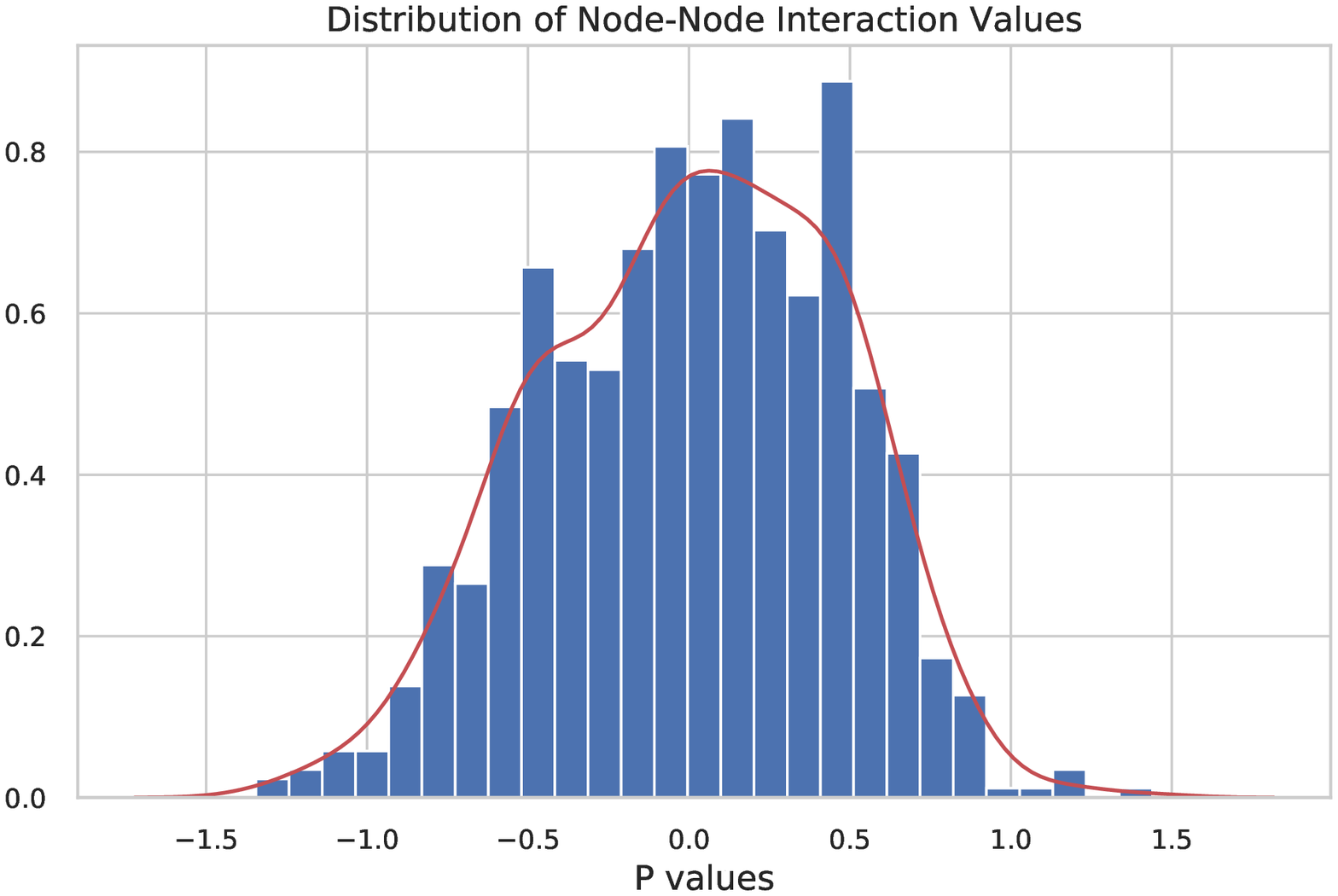}
\end{minipage}
\caption{The heatmap (top) and distribution(bottom) of $\mP$ values (scaled by 2).}
\label{fig4}
\end{figure}

$\mP$ learns the node--node interactions within a graph structure. Fig. \ref{fig4} shows a heatmap of $\mP$ from last attentional layer and first head of resnet \#3 trained using 99/1 data split from CHAMPS dataset. Although the adjacency matrix was symmetric for this dataset, $\mP$ is not a symmetric matrix. The node--node relationships in this map generalizes the data between graph nodes based on their position in the adjacency matrix. The power values (scaled by 2, see eq. \ref{eq2}) vary in a range of $[-1.345, 1.440]$ in the heatmap and the distributions of these values is gaussian-like. The range shows that for some nodes the interaction scales linearly with distance $r^{2|P_{ij}|}$ rather than $1/r^{2|P_{ij}|}$. This may seem unphysical and counterintuitive, however, such effects of linear scaling of interaction force with distance are observed as quark confinement in the nature and accounted in analysis of quarkonium using the potential expression\cite{Eichten80, Sumino03}:

\begin{equation}
a r+\dfrac b r +V_0
\label{eq11}
\end{equation}

The graph attention model learns potential terms similar to eq. \ref{eq11} for the CHAMPS dataset. In the heatmap, graph node pairs can be classified as having a confinement interaction ($P_{ij} < 0$), coulomb interaction ($P_{ij} > 0$) or distance independent interaction ($P_{ij} = 0$). The distribution of $2P_{ij}$ values in Fig. \ref{fig4} has nearly bell curve shape for resnet \#3 with a mean value of 0.007 and standard deviation of 0.456.

\begin{figure}[!h]
\centering
\begin{minipage}[b]{0.8\textwidth}
 \centering
 \includegraphics[width=\textwidth]{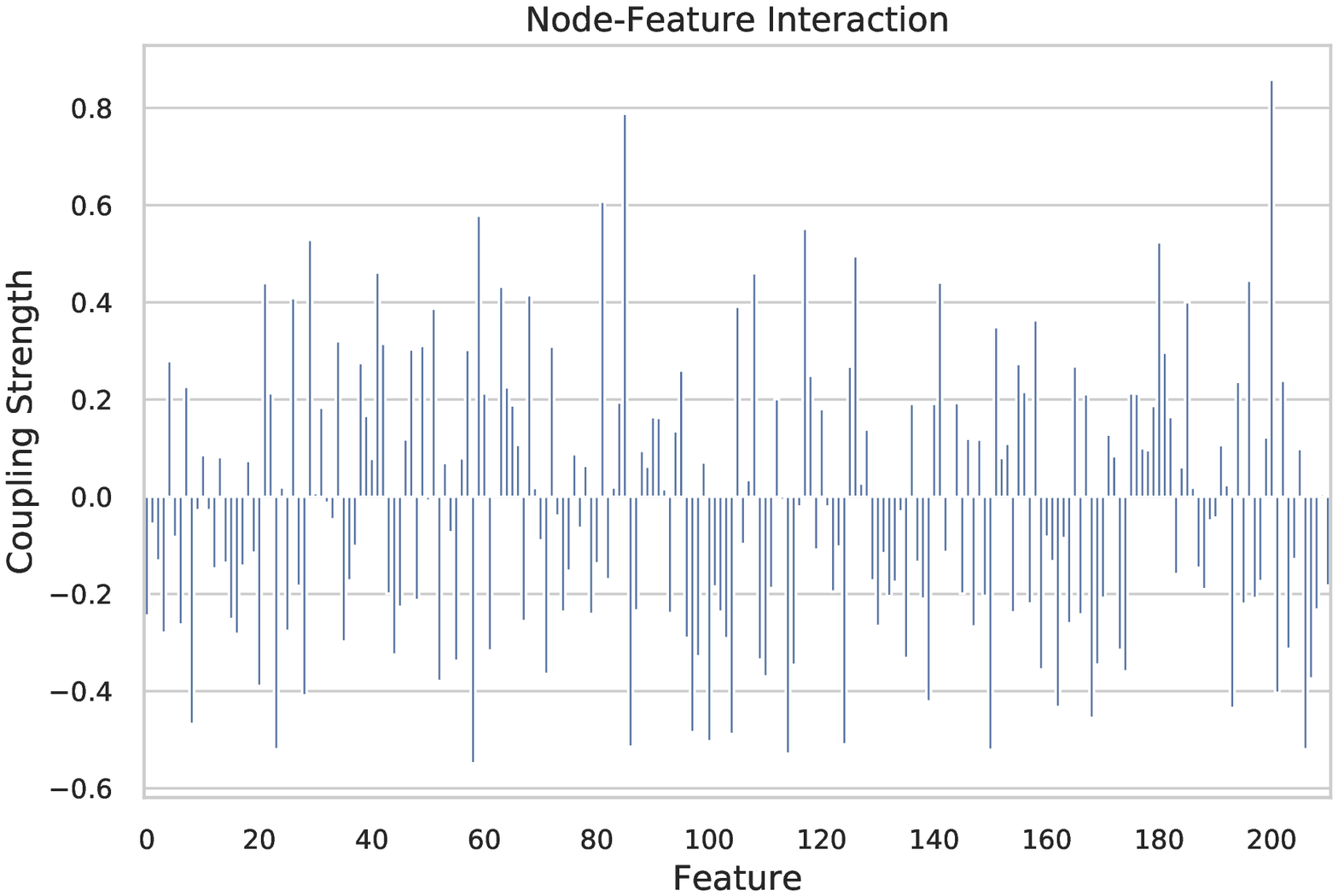}
\end{minipage}
\hfill
\hfill
\begin{minipage}[b]{0.8\textwidth}
 \centering
 \includegraphics[width=\textwidth]{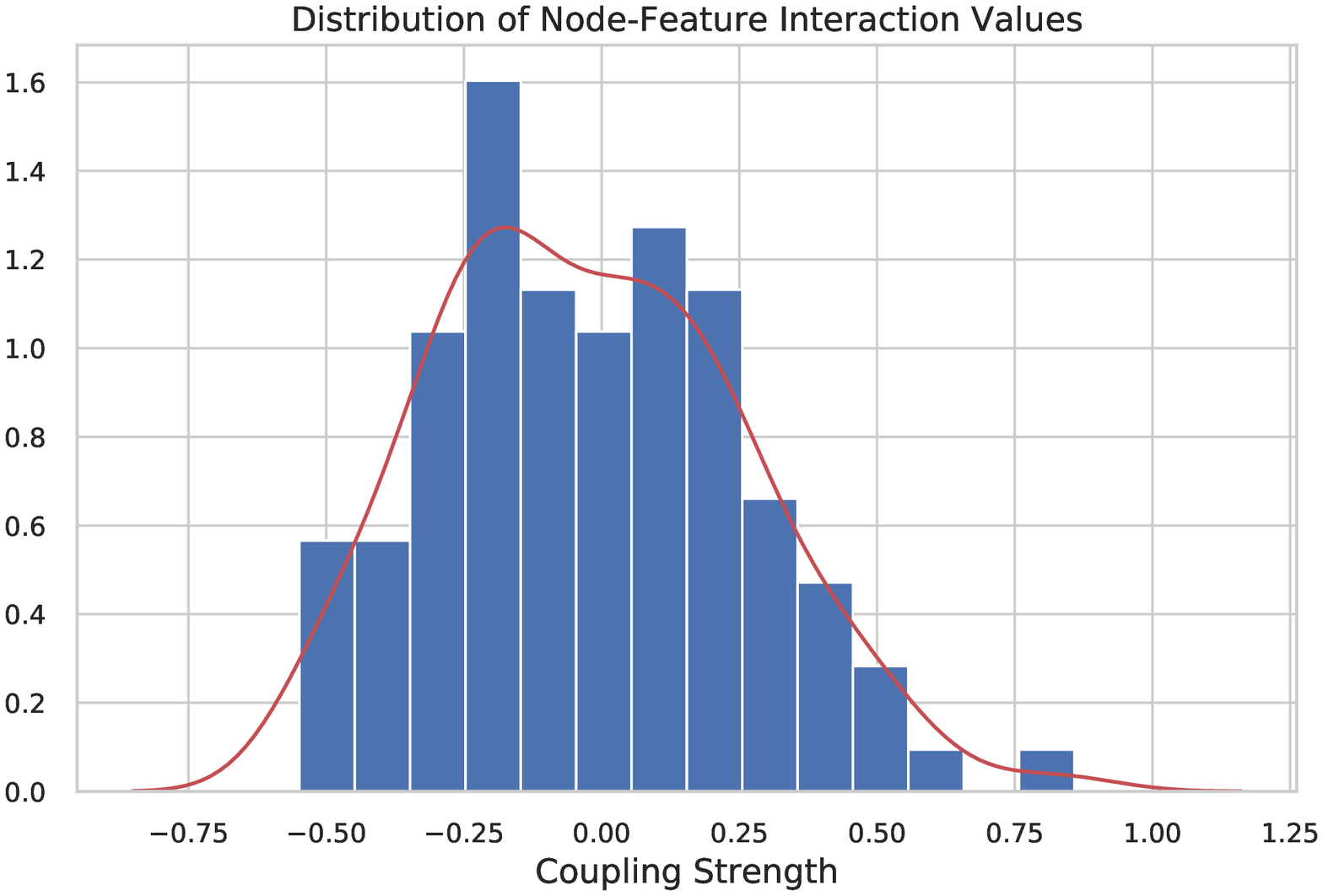}
\end{minipage}
\caption{The barplot (top) and distribution (bottom) of $\va$ values.}
\label{fig5}
\end{figure}

Fig. \ref{fig5} shows a barplot of a single columns from $\va$ for node 0 and a set of 211 features for CHAMPS data. Although the columns of input and output are sparse, column of $\va$ is not, indicating complex relationships between a single node and features. For CHAMPS data, this $\va$ column shows a range of $[-0.549, 0.858]$ for interaction strength for each feature. The distribution of $a_{ij}$ also has close to a bell curve shape with mean value of 0.025 and standard deviation of 0.277.

\begin{figure}[!h]
\centering
\begin{minipage}[b]{0.7\textwidth}
 \centering
 \includegraphics[width=\textwidth]{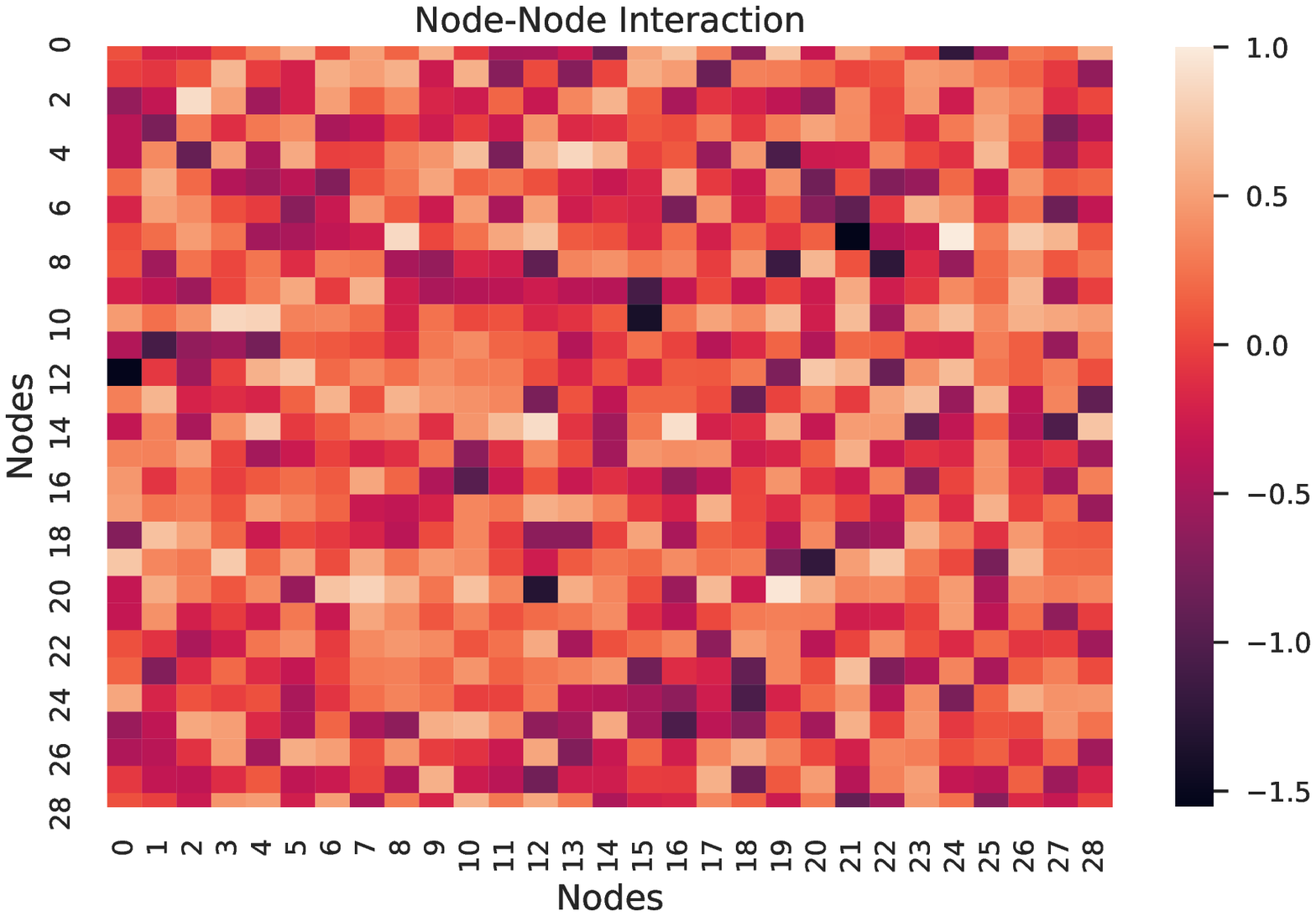}
\end{minipage}
\hfill
\begin{minipage}[b]{0.7\textwidth}
 \centering
 \includegraphics[width=\textwidth]{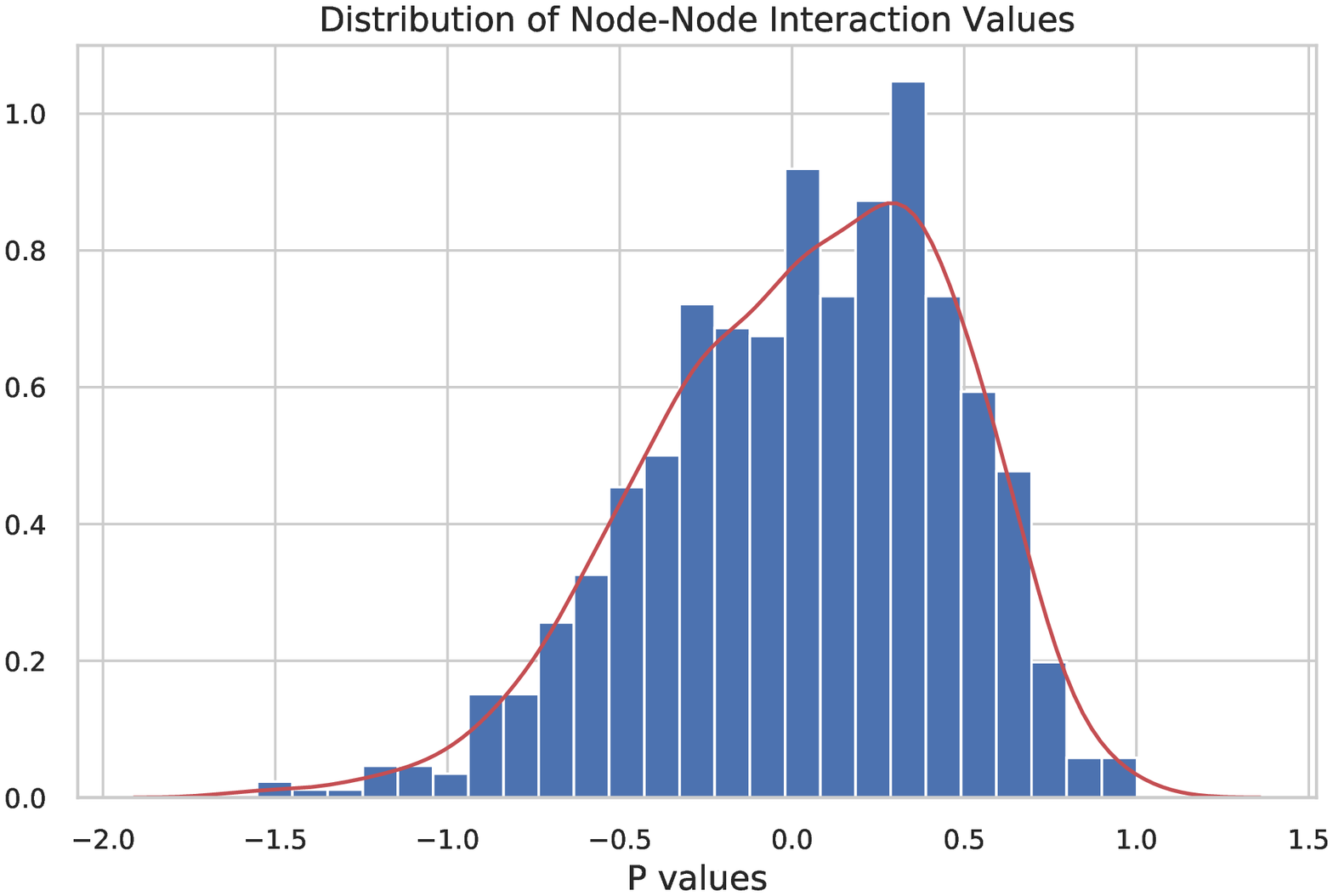}
\end{minipage}
\caption{The heatmap (top) and distribution (bottom) of $\mP$ values for resnet \#7 (scaled by 2).}
\label{fig8}
\end{figure}

\begin{figure}[!h]
\centering
\begin{minipage}[b]{0.8\textwidth}
 \centering
 \includegraphics[width=\textwidth]{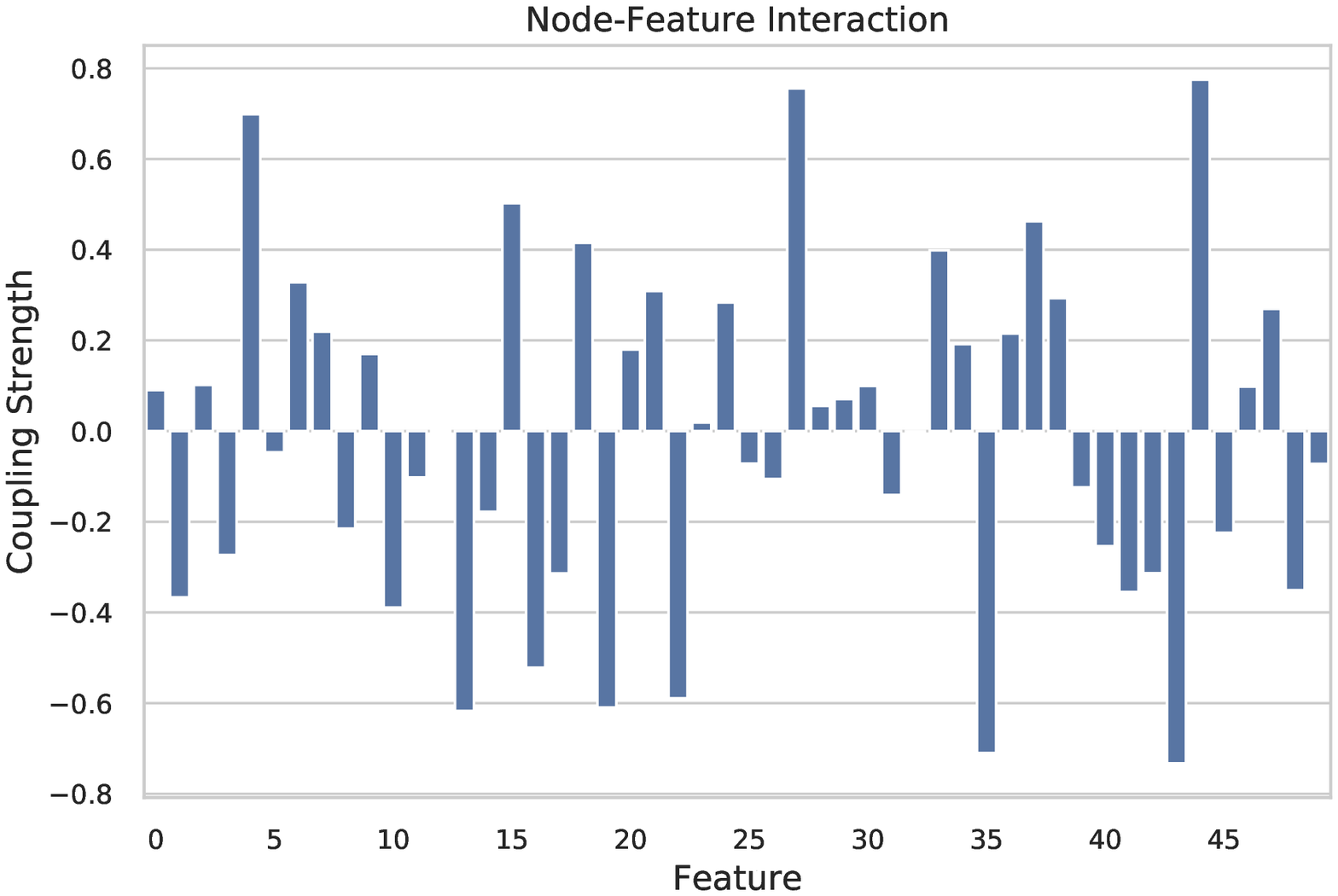}
\end{minipage}
\hfill
\hfill
\begin{minipage}[b]{0.8\textwidth}
 \centering
 \includegraphics[width=\textwidth]{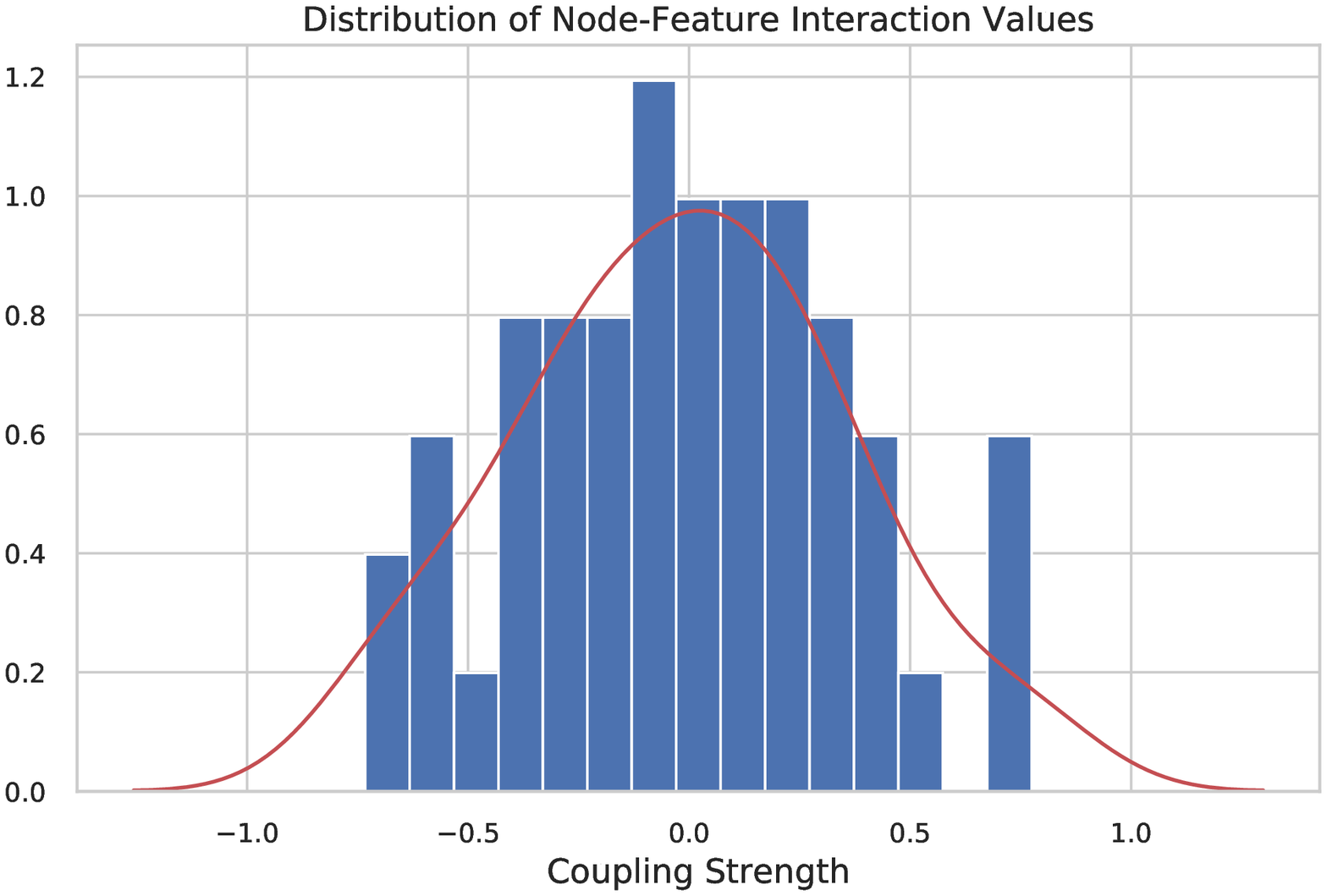}
\end{minipage}
\caption{The barplot (top) and distribution (bottom) of $\va$  values for resnet \#7.}
\label{fig9}
\end{figure}

When we evaluated deeper models, we reduced the hidden layer feature size to form deeper models. This classification of hidden layers from learnable attention variables makes it possible to compare two models trained on same data on a layer-by-layer basis. Fig. \ref{fig8} and \ref{fig9} shows the heatmap, barplot and distributions of resnet \#7 trained with 99/1 data split. The range, mean and standard deviation of $\va$ and $\mP$ are summarized in Table \ref{table31} for last layer and first head. The mean and standard deviation are comparable for $\mP$ with resnet \#3, $\va$ distribution has a larger standard deviation for 50 hidden features as compared to 211 hidden features in the last layer for same node. This suggests that deeper models with smaller hidden feature sizes have a more uniform distribution of node--feature interaction. 

The ability to compare single attention layers among very different models provides another tool for designing hidden layers with optimum node--node and node--feature interaction characteristics for a given dataset. Therefore, a screened coulomb potential attention model can be used to extract an empirical standard model of graph structure of a  dataset from the model or to find the optimum representations for nodes and hidden features of a model from the dataset. These two processes can be applied iteratively to optimize the model representation and to interpret the graph structure more accurately. 

\begin{table}
\caption{$\mP$ and $\va$ distribution of single attentional layer.}
\label{table31}
\resizebox{\textwidth}{!}{
\begin{tabular}{c c c c c c c c c c }
\hline
Model Name & \multicolumn{4}{c} {$P_{ij}$}  &  \multicolumn{4}{c} {$a_{ij}$}  \\
\hline
  & Lower Bound & Upper Bound & Mean & Std. Dev. & Lower Bound & Upper Bound & Mean & Std. Dev.\\
\hline
Resnet \#3 &  -1.345 & 1.440 & 0.008 & 0.456 &  -0.549 & 0.858 & 0.025 & 0.277  \\ 
\hline 
Resnet \#7 & -1.553 & 1.002 & 0.0308 & 0.432 & -0.733 & 0.775 & -0.013 & 0.364 \\
\hline 
\end{tabular} }
\end{table}

\clearpage

\section{Conclusions}

In this paper, we presented an efficient graph attention model that interprets the relationship between a node and its neighbors by learning the coupling strength and range of interaction between nodes using an attention form inspired from definition of screened Coulomb potential. The new attention mechanism employs a weighted adjacency matrix, learnable power variables $\va$ and $\mP$ that learns the interaction strength and range from dataset. CHAMPS dataset was used to characterize the capabilities of a variety of implementations of CoulGAT framework after preprocessing the dataset to have each molecule represented as a graph. Stable plain and resnet graph models having up to 140 layers and 10 heads were demonstrated and characterized with this attention mechanism. 

The learnable parameters $(\va, \mP)$ present a way to quantify the node--node and node-feature interactions from the hidden layer to interpret graph structure of the training dataset through a simple, empirical standard model. The learnable parameters also provide valuable information about the representation power of the hidden attentional layer which can be compared among different models and optimized using statistical distribution of node and feature interactions.

As future work, evaluation of this attention mechanism with different transductive and inductive graph datasets, better input embeddings and other high capacity models such as transformers can provide more insights on the capabilities of the screened Coulomb potential as a tool for interpretability of both hidden layers of a graph model and graph structure of a dataset.

\subsubsection*{Acknowledgments}

I would like to acknowledge numerous contributions of machine learning research community on graph networks in recent years. I thank my parents for their support and patience. This research was conducted independently without support from a grant or corporation.

\bibliography{coulgat_paper}

\bibliographystyle{ieeetr}

\pagebreak

\appendix

\section{MSE LOSS CURVES}

\begin{figure}[!hbt]
\centering
\begin{minipage}[b]{0.7\textwidth}
 \includegraphics[width=\textwidth]{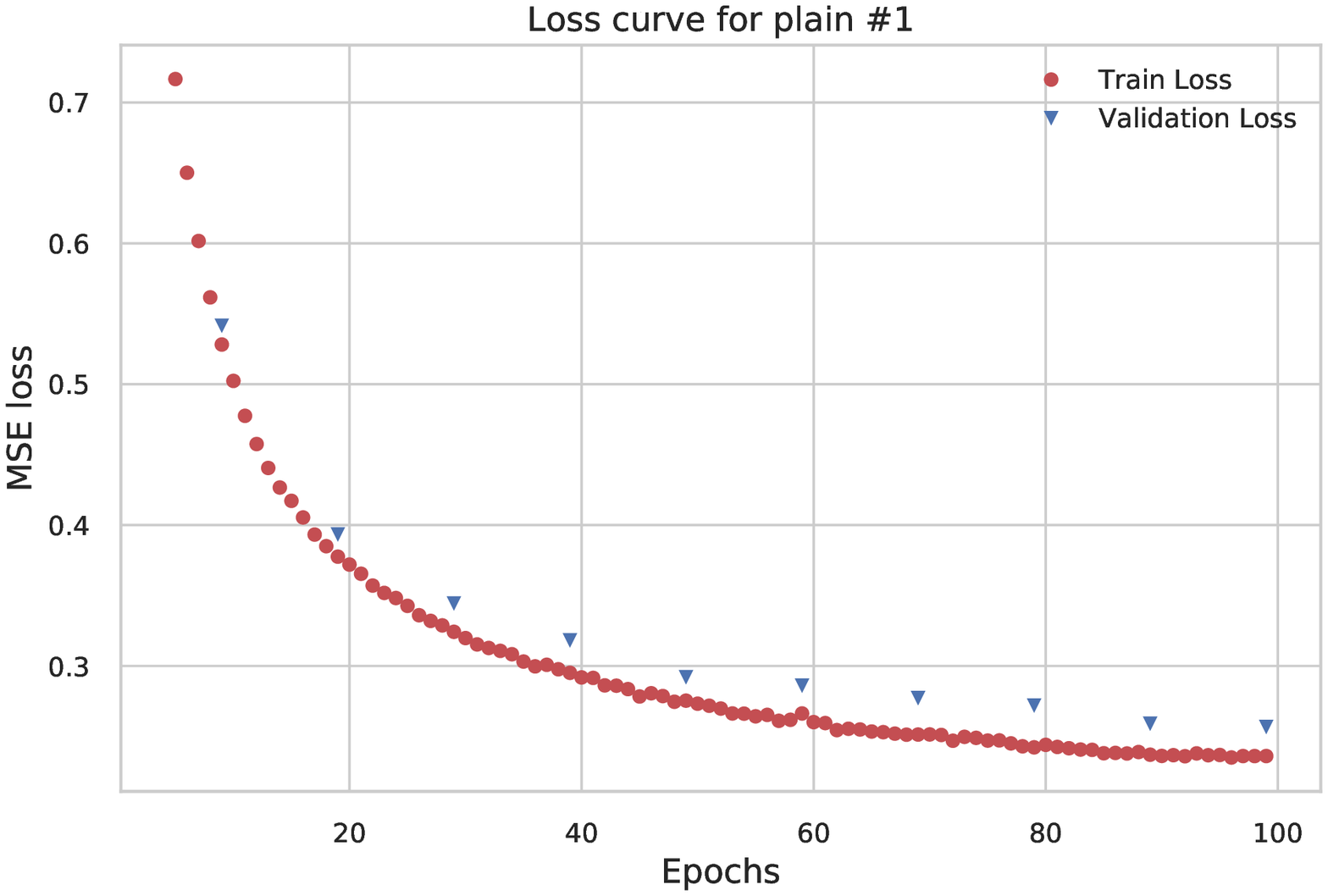}
\end{minipage}
\hfill
\begin{minipage}[b]{0.7\textwidth}
 \includegraphics[width=\textwidth]{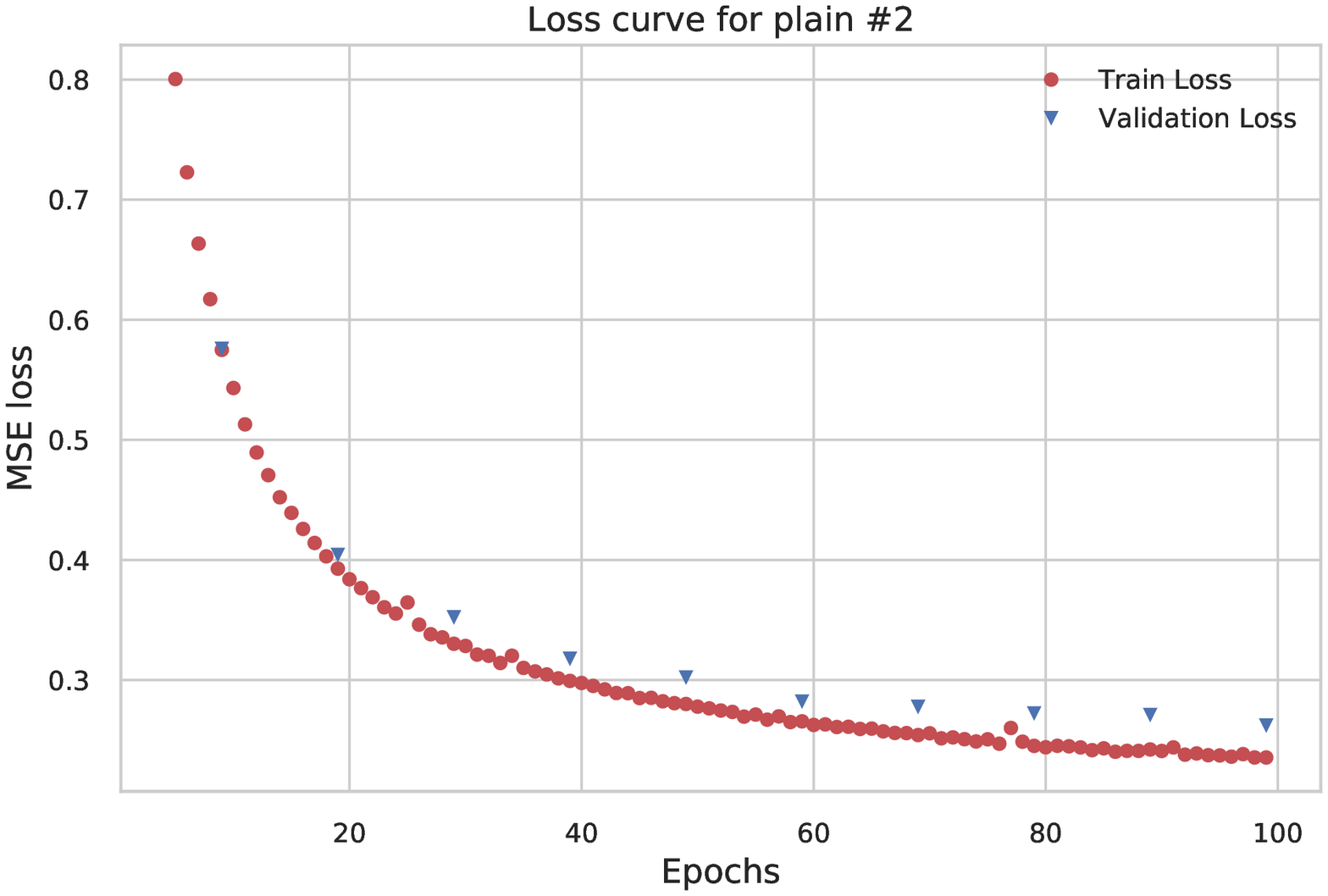}
\end{minipage}
\hfill
\begin{minipage}[b]{0.7\textwidth}
 \includegraphics[width=\textwidth]{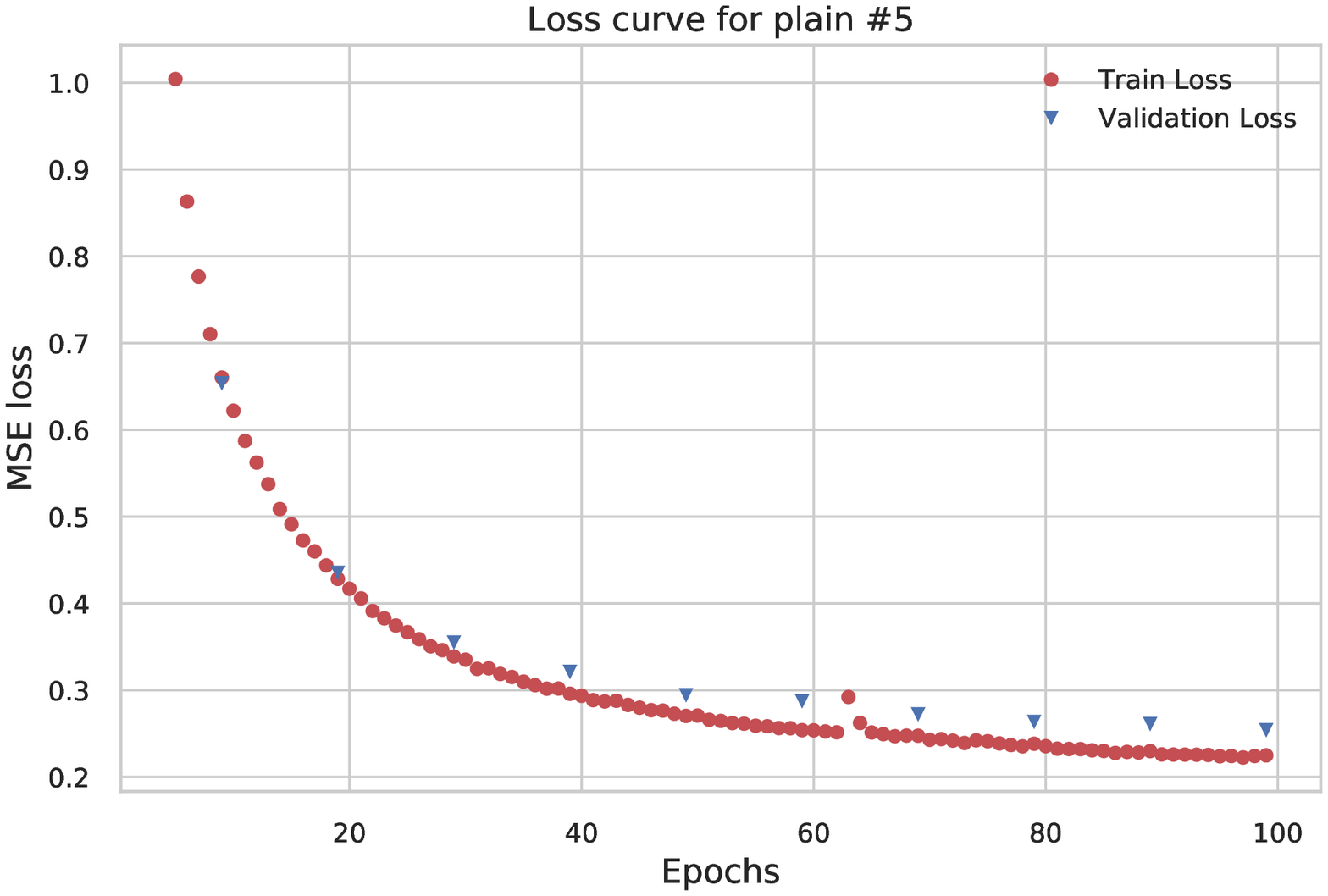}
\end{minipage}
\end{figure}

\begin{figure}[!hbt]
\centering
\begin{minipage}[b]{0.7\textwidth}
 \includegraphics[width=\textwidth]{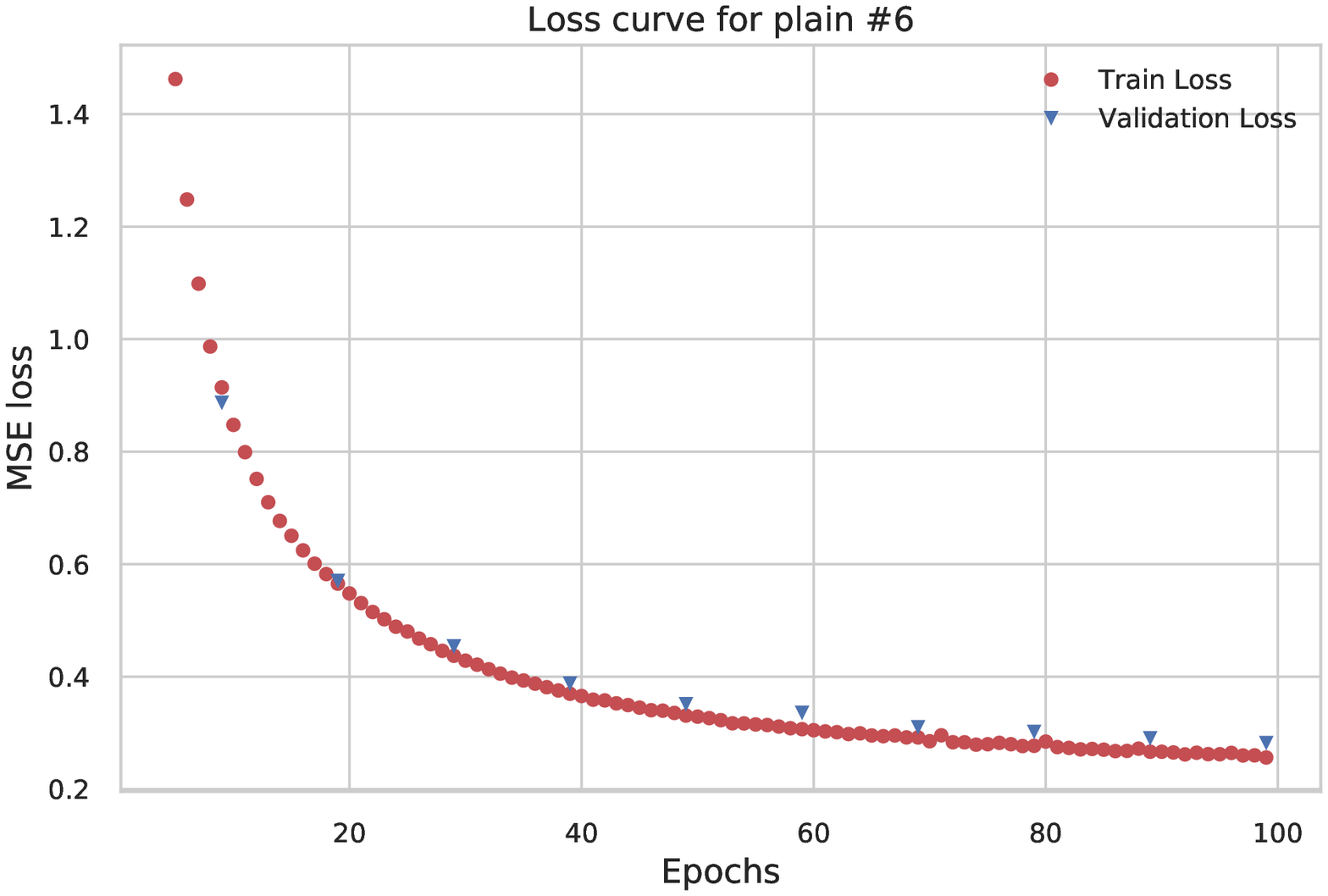}
\end{minipage}
\hfill
\begin{minipage}[b]{0.7\textwidth}
 \includegraphics[width=\textwidth]{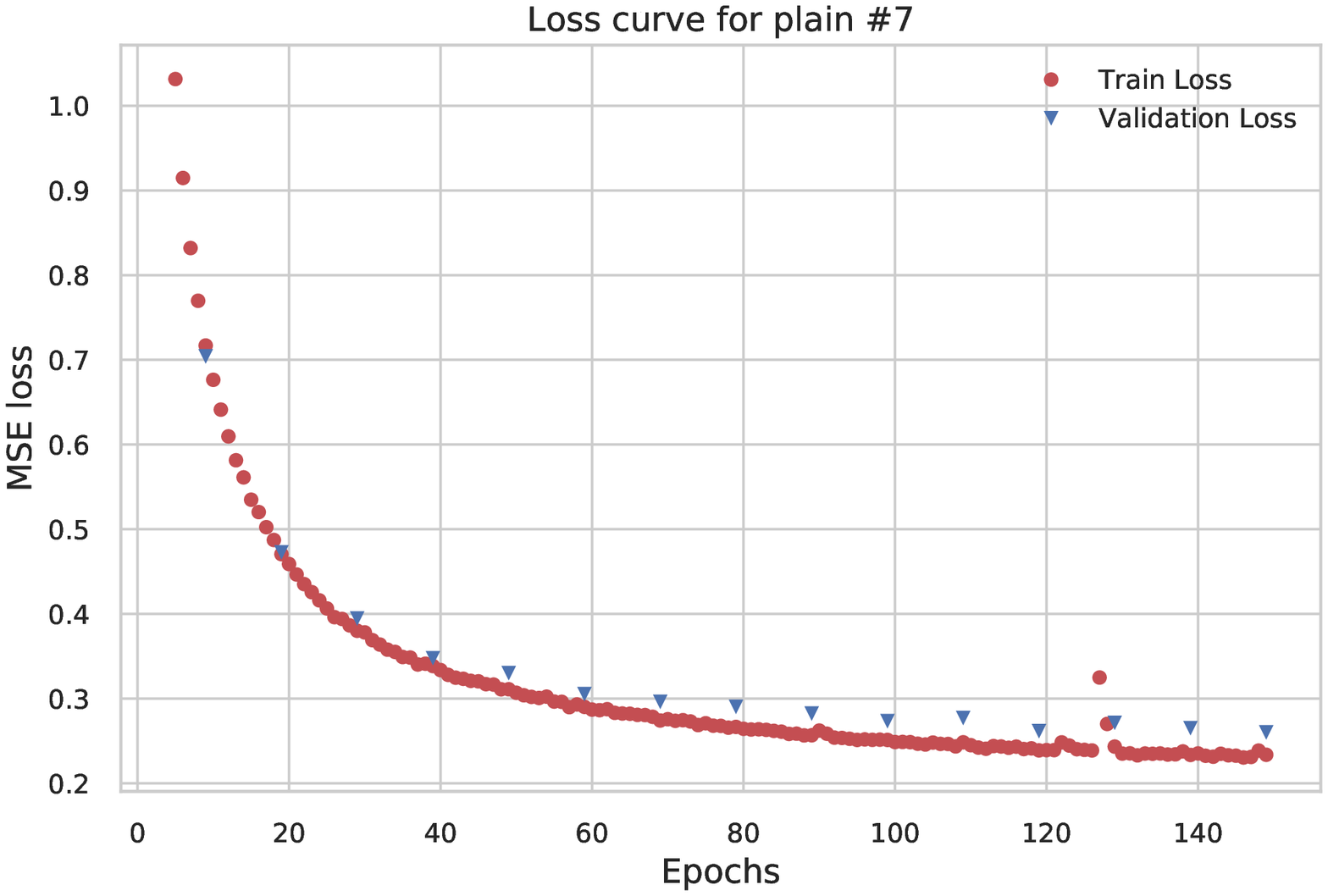}
\end{minipage}
\hfill
\begin{minipage}[b]{0.7\textwidth}
 \includegraphics[width=\textwidth]{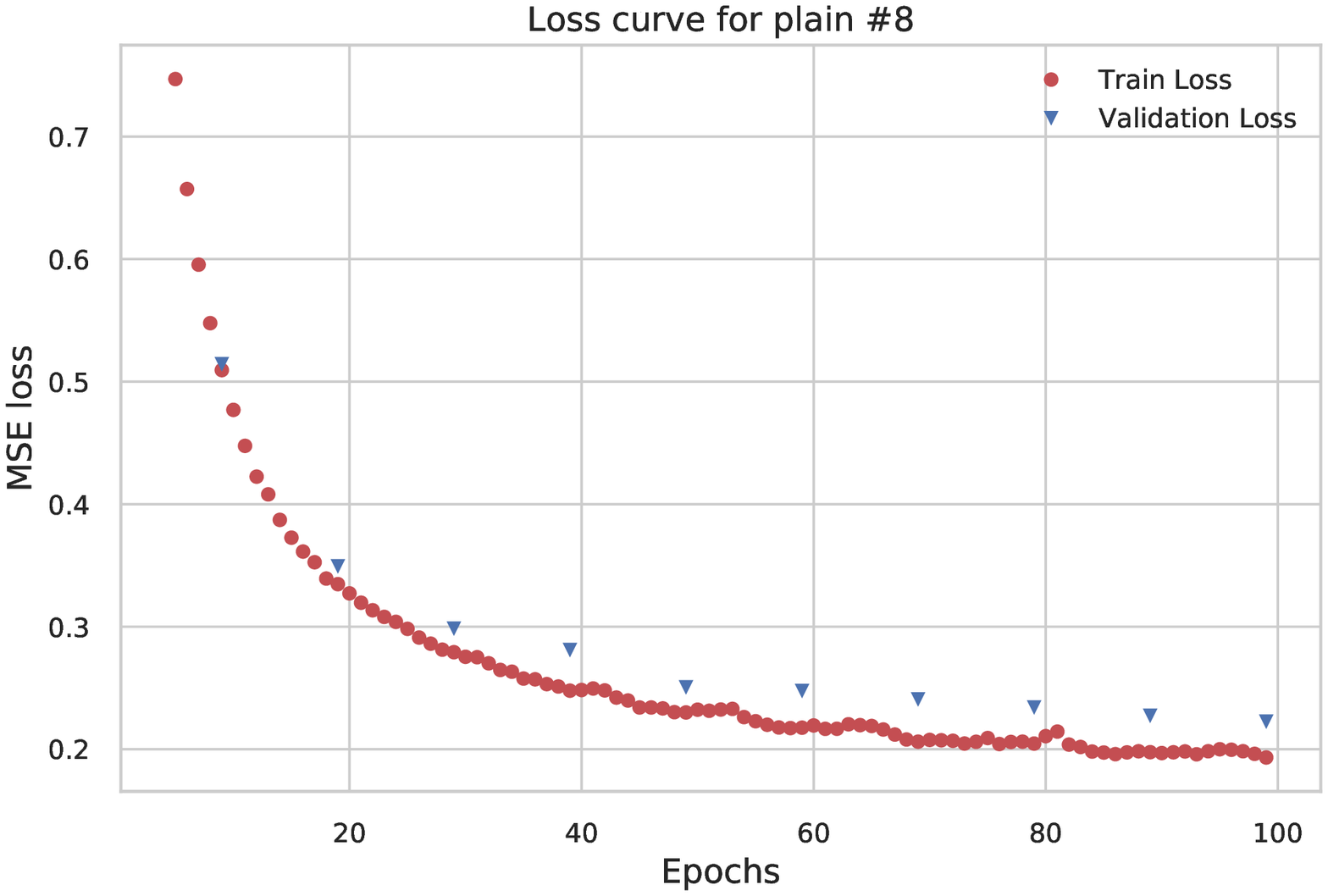}
\end{minipage}
\label{app1_mse_plain}
\end{figure}

\begin{figure}[!hbt]
\centering
\begin{minipage}[b]{0.7\textwidth}
 \includegraphics[width=\textwidth]{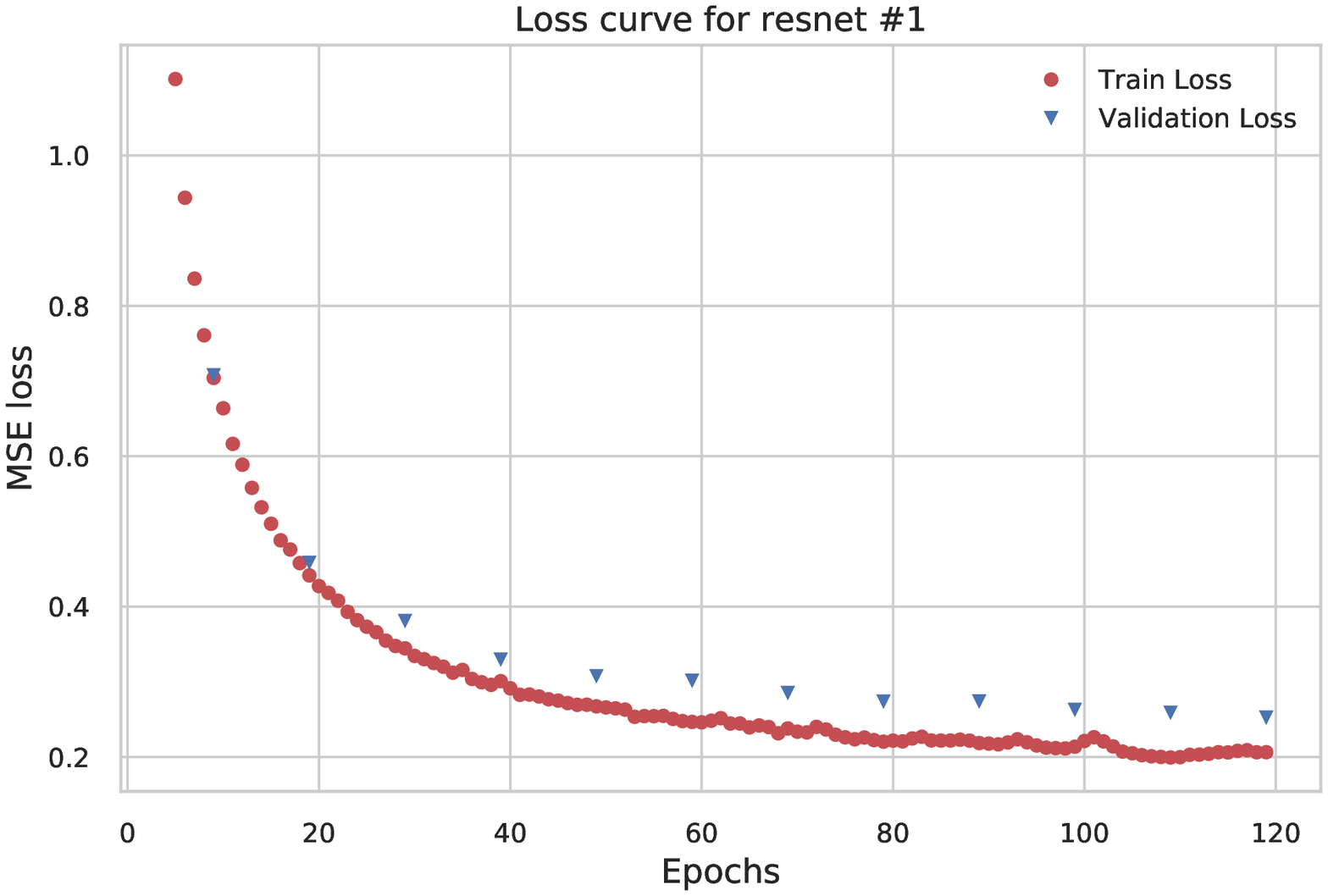}
\end{minipage}
\hfill
\begin{minipage}[b]{0.7\textwidth}
 \includegraphics[width=\textwidth]{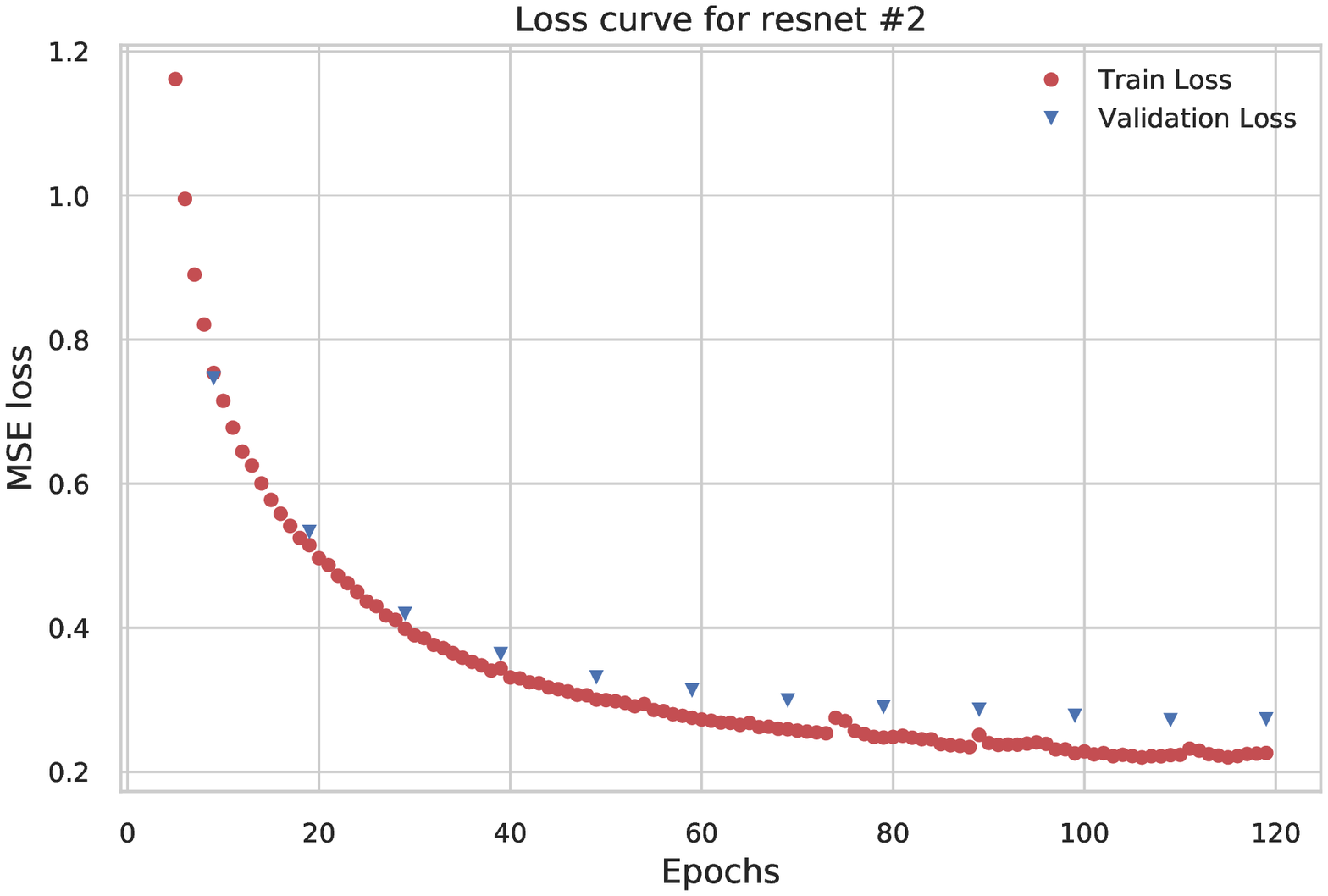}
\end{minipage}
\hfill
\begin{minipage}[b]{0.7\textwidth}
 \includegraphics[width=\textwidth]{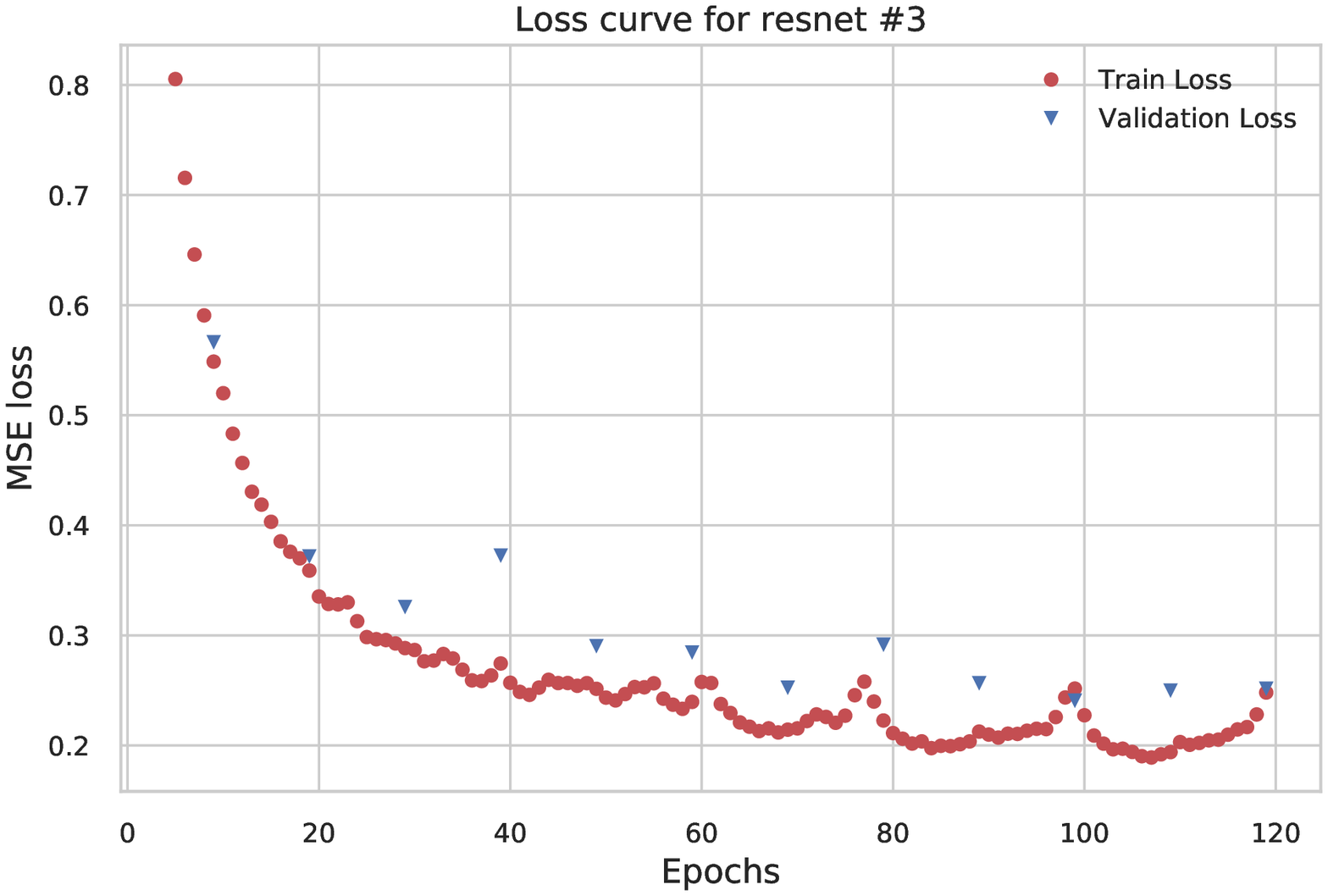}
\end{minipage}
\end{figure}

\begin{figure}[!hbt]
\centering
\begin{minipage}[b]{0.7\textwidth}
 \includegraphics[width=\textwidth]{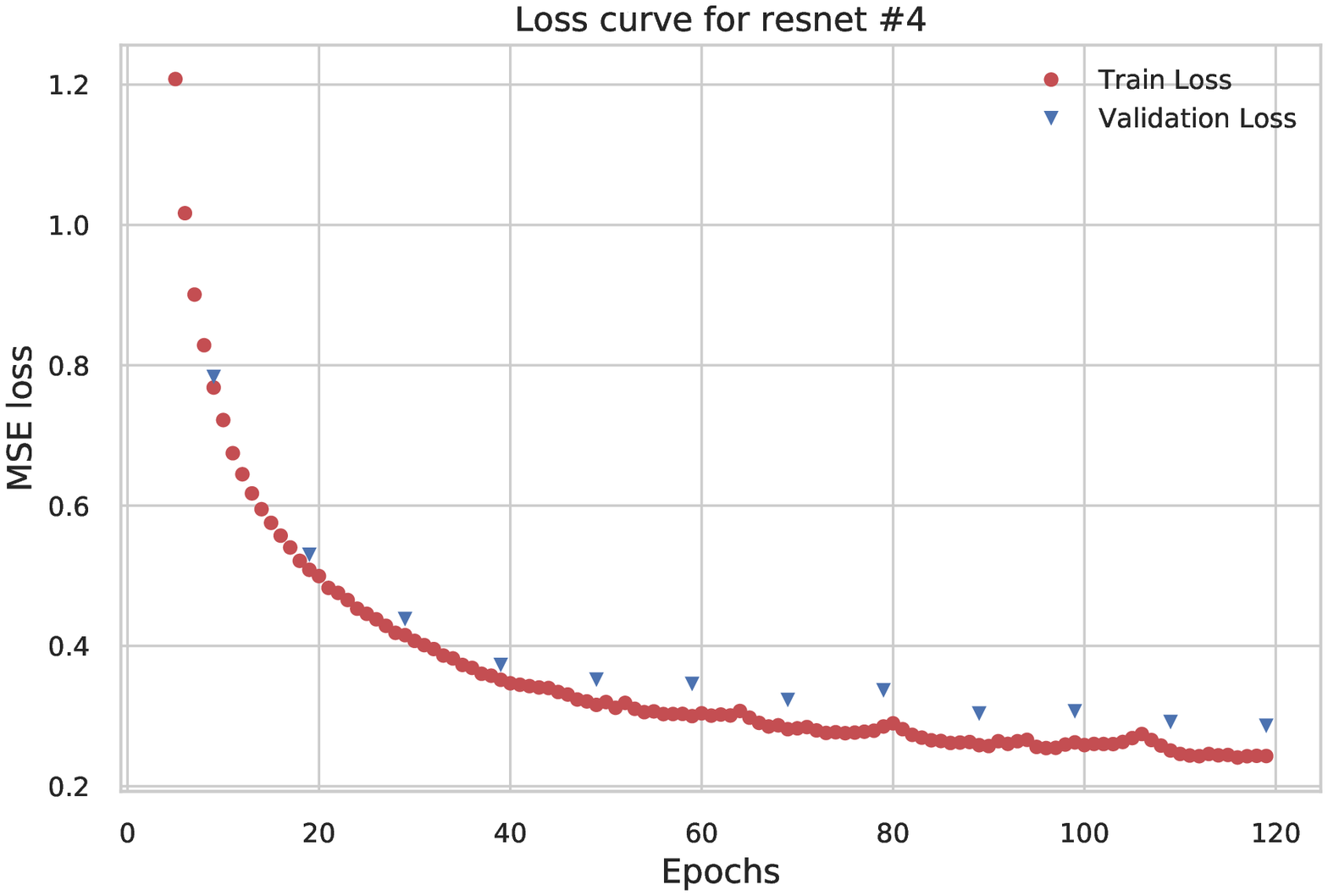}
\end{minipage}
\hfill
\begin{minipage}[b]{0.7\textwidth}
 \includegraphics[width=\textwidth]{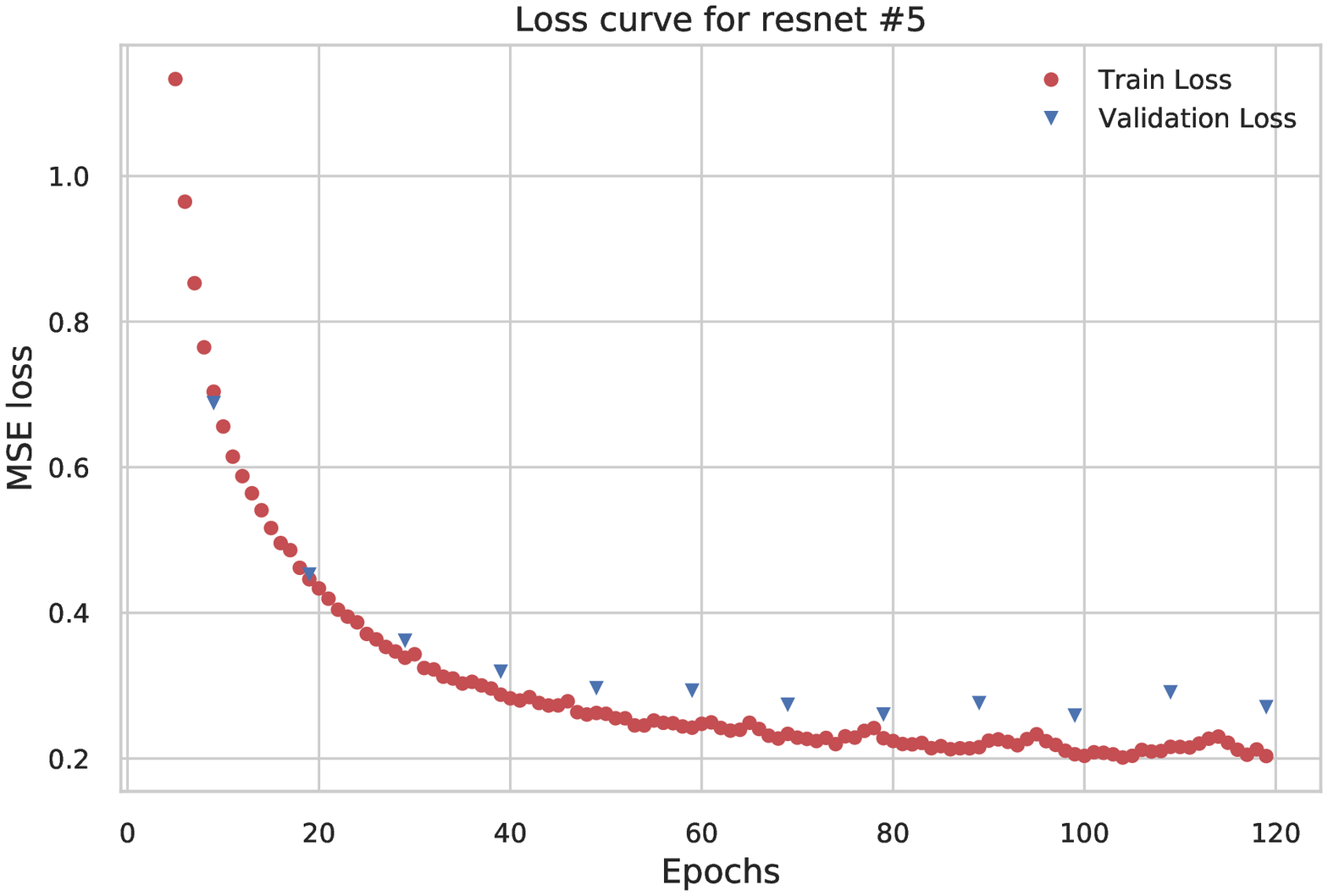}
\end{minipage}
\label{app1_mse_plain}
\end{figure}

\clearpage
\section{SCCLMAE LOSS CURVES}

\begin{figure}[!hbt]
\centering
\begin{minipage}[b]{0.7\textwidth}
 \includegraphics[width=\textwidth]{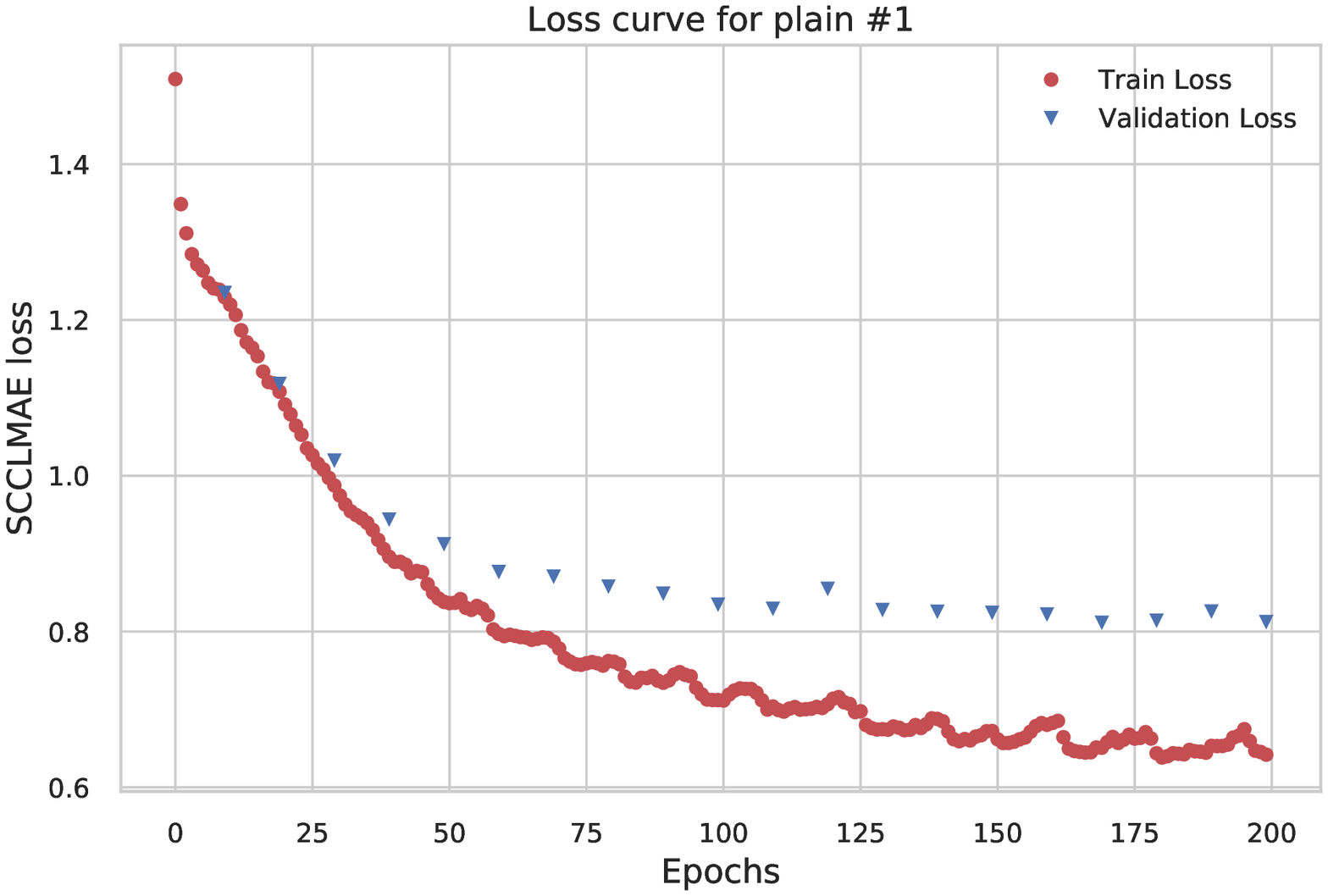}
\end{minipage}
\hfill
\begin{minipage}[b]{0.7\textwidth}
 \includegraphics[width=\textwidth]{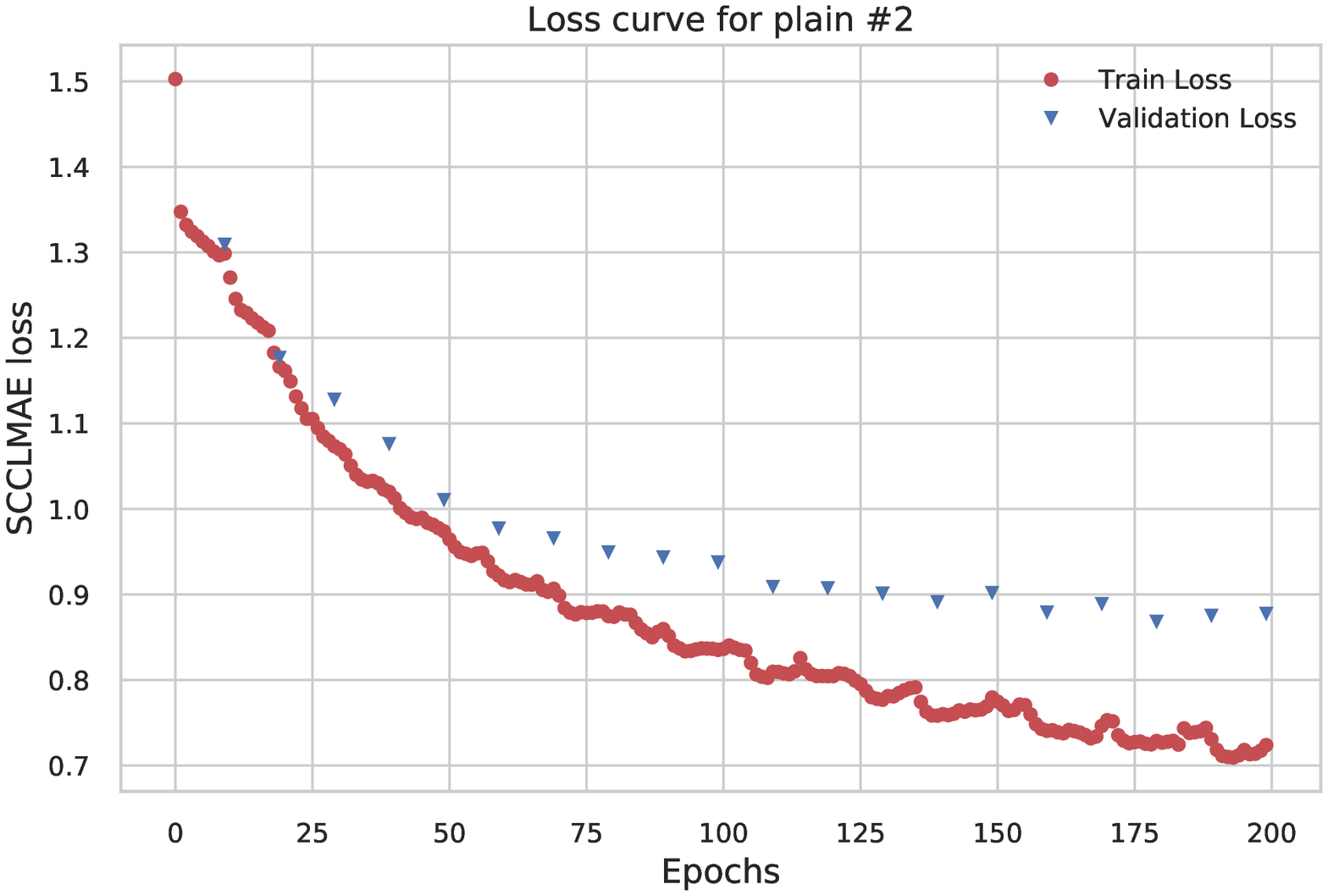}
\end{minipage}
\hfill
\begin{minipage}[b]{0.7\textwidth}
 \includegraphics[width=\textwidth]{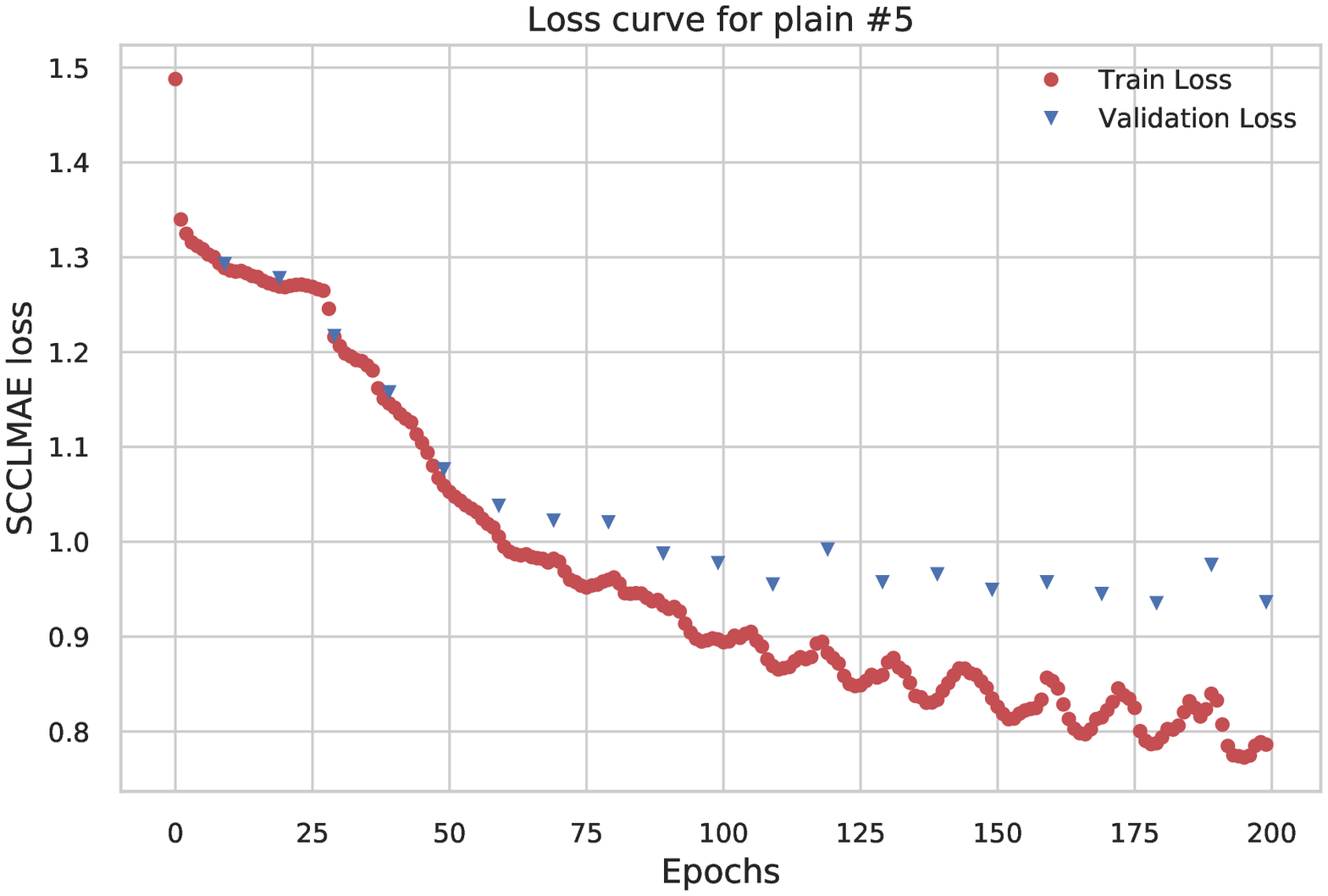}
\end{minipage}
\end{figure}

\begin{figure}[!hbt]
\centering
\begin{minipage}[b]{0.7\textwidth}
 \includegraphics[width=\textwidth]{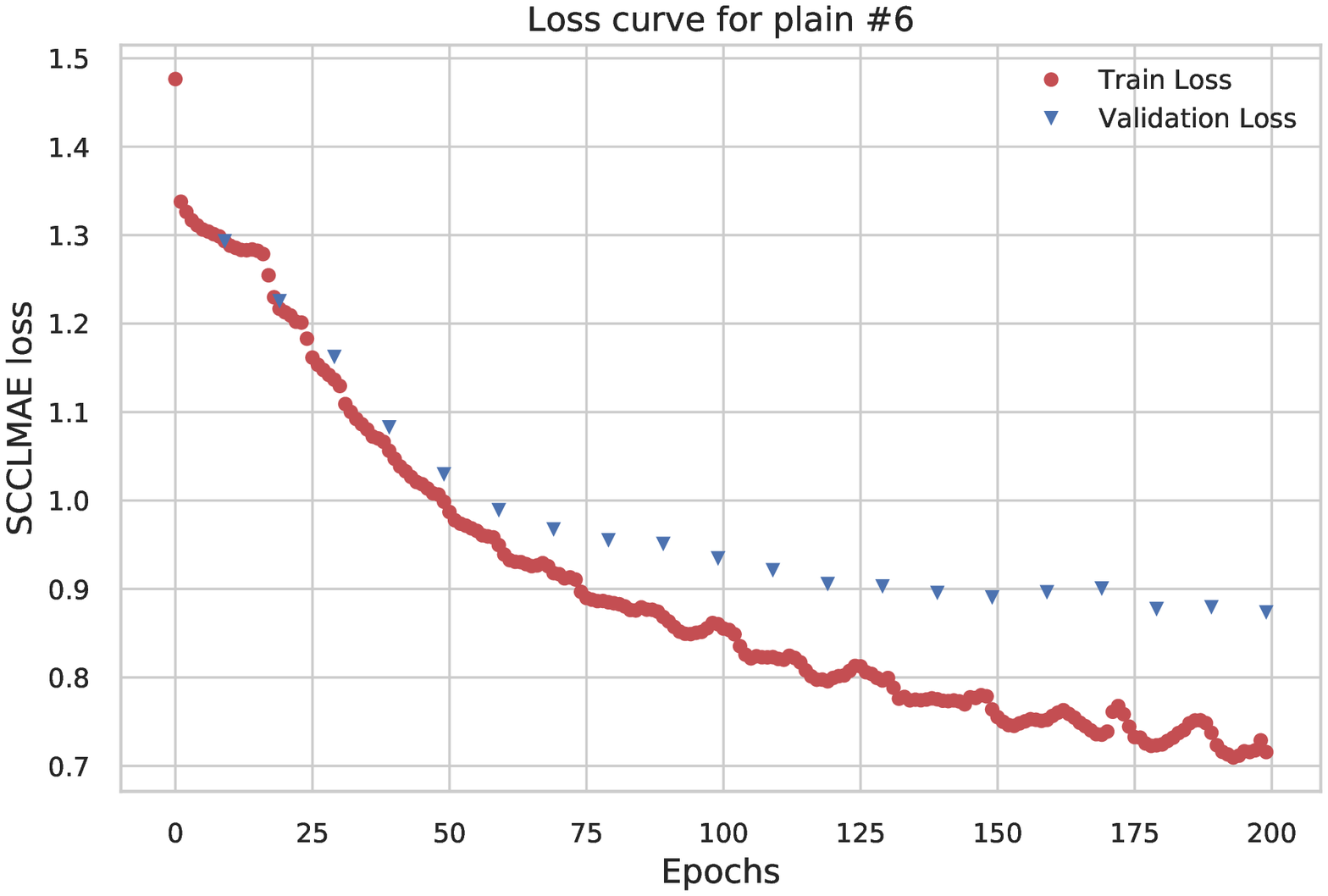}
\end{minipage}
\label{app1_scclmae_plain}
\end{figure}

\begin{figure}[!hbt]
\centering
\begin{minipage}[b]{0.7\textwidth}
 \includegraphics[width=\textwidth]{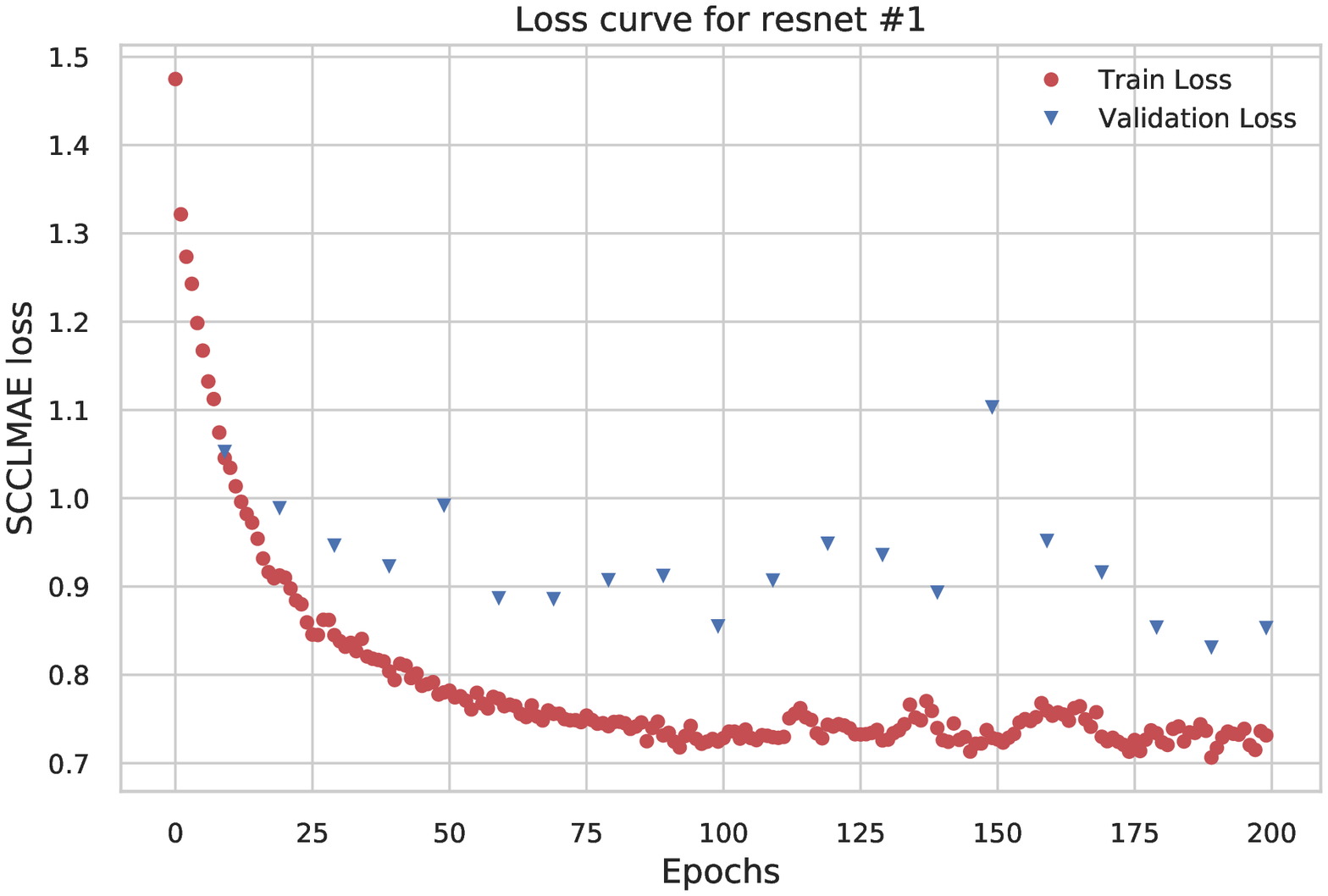}
\end{minipage}
\hfill
\begin{minipage}[b]{0.7\textwidth}
 \includegraphics[width=\textwidth]{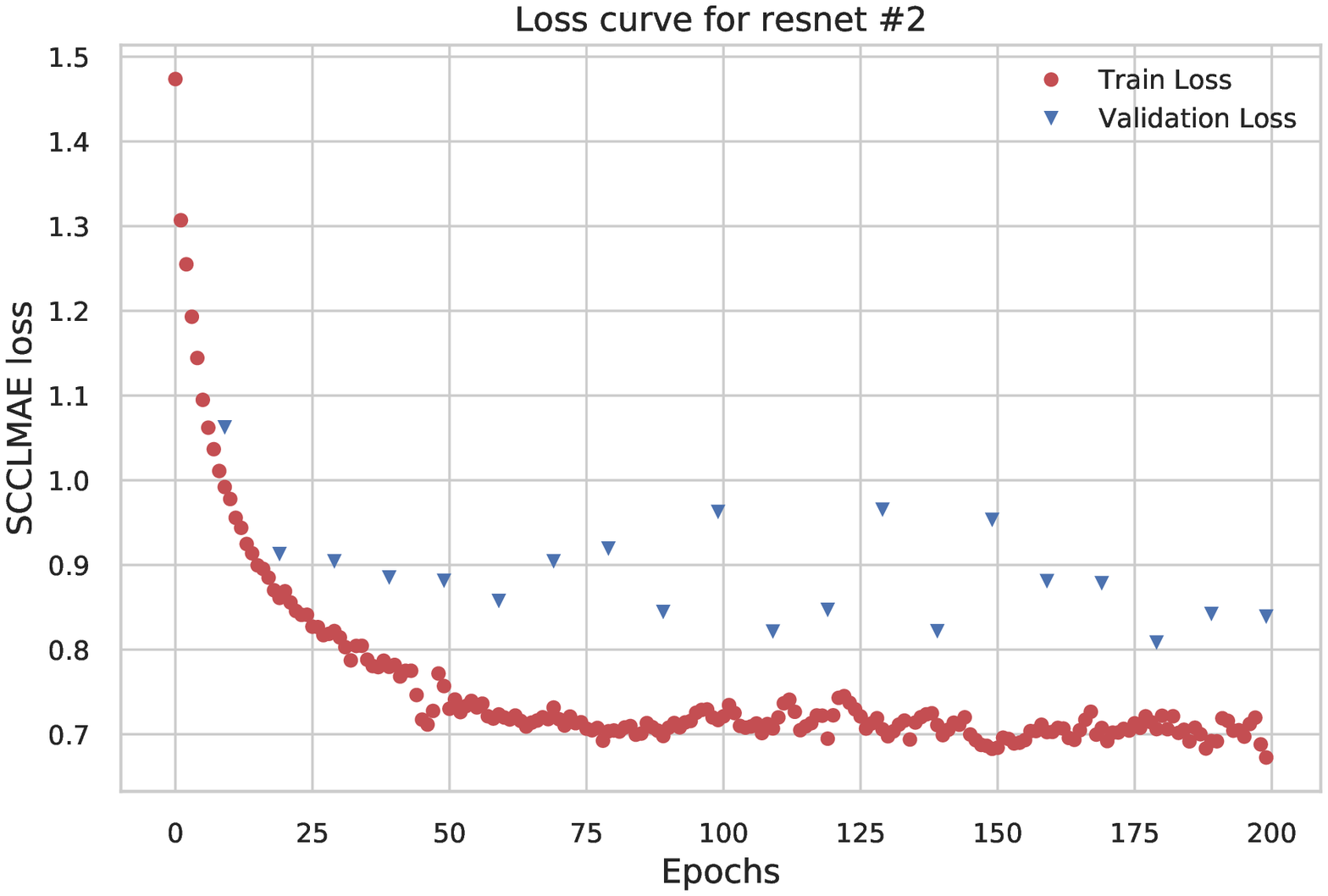}
\end{minipage}
\end{figure}
\begin{figure}[!hbt]
\centering
\begin{minipage}[b]{0.7\textwidth}
 \includegraphics[width=\textwidth]{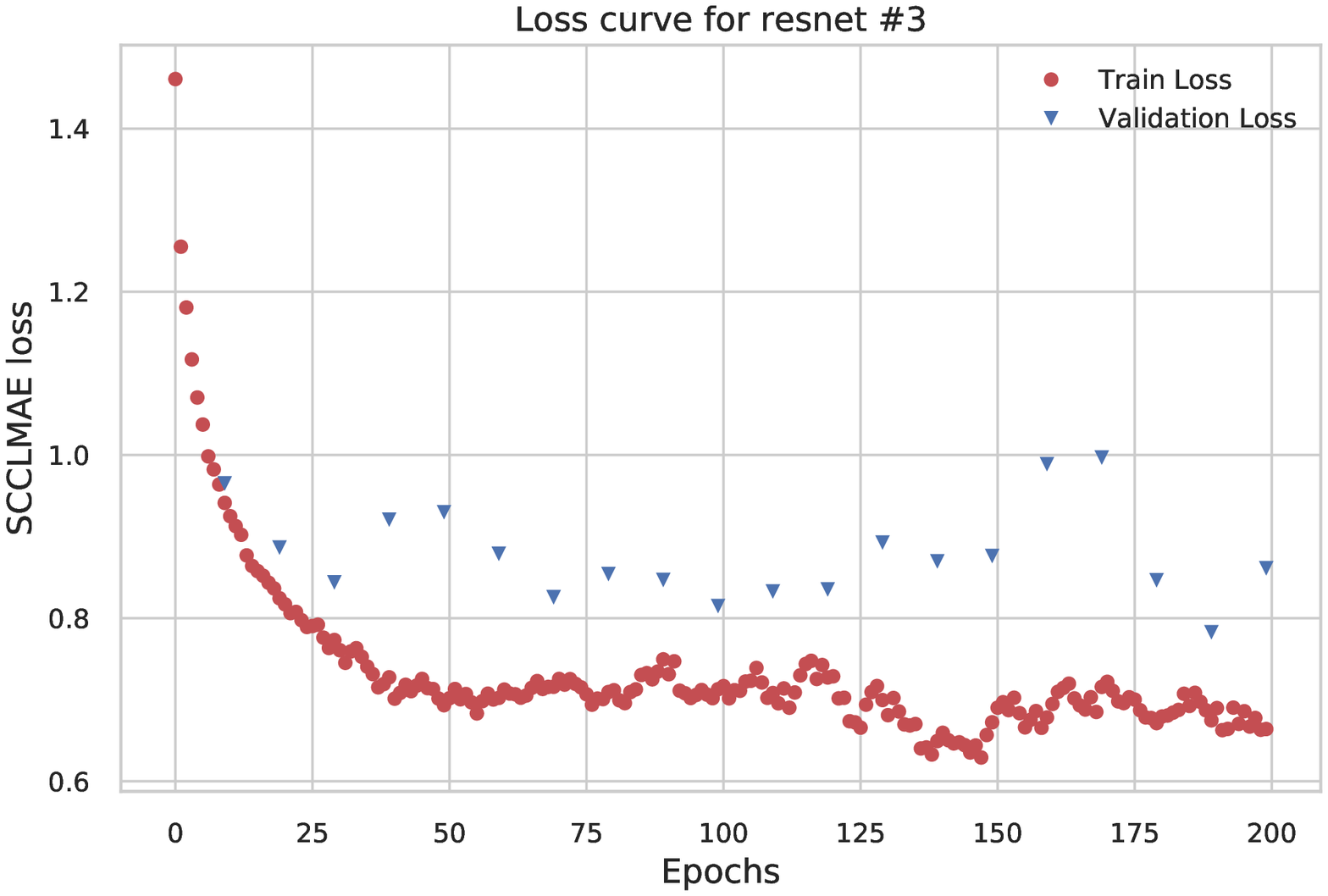}
\end{minipage}
\hfill
\begin{minipage}[b]{0.7\textwidth}
 \includegraphics[width=\textwidth]{scclmae_resnet4_pred.eps}
\end{minipage}
\hfill
\begin{minipage}[b]{0.7\textwidth}
 \includegraphics[width=\textwidth]{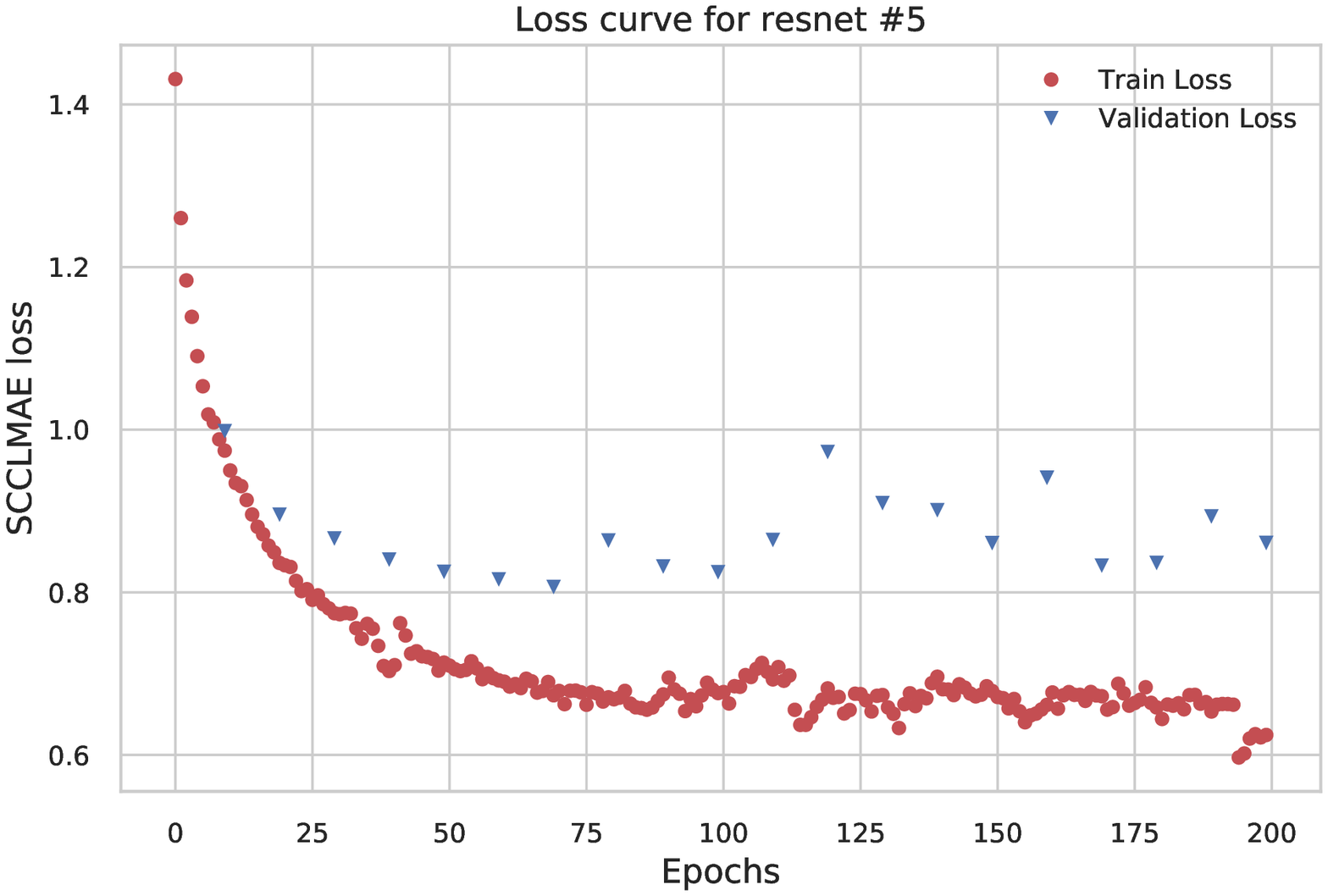}
\end{minipage}
\label{app1_scclmae_resnet}
\end{figure}

\end{document}